\PassOptionsToPackage{table}{xcolor}
\documentclass{article}
\usepackage{arxiv}

\input{preamble}

\title{ProClosure: Hierarchical Room-Object Assignment using Progressive Boundary Closure from Monocular Video
\thanks{Code: \url{https://github.com/ClarityLab-Org/ProClosure}}
}

\author{Vinoth Kumar Muthuraj, Soumyadeep Banik, Kushal Sharma, Hardik Jain \\
Clarity Lab, Department of CSE, IIT Jodhpur \\
  \texttt{m25ai1130@alumni.iitj.ac.in, \{soumyadeep, m25csa016, hardik.jain\}@iitj.ac.in} \\
}

\begin{document}
\maketitle

\begin{abstract}
A 3D scene graph groups objects into rooms. When a robot is asked to fetch an object from the kitchen, that grouping is what tells it where to look, and an object recorded in the wrong room is not retrievable by a query naming the correct room. We introduce \textit{Progressive Boundary Closure} (ProClosure), which recovers the room layer from a single monocular RGB video. A SLAM front end and an open-vocabulary segmenter supply a structural point cloud, a camera trajectory and object tracks; the cloud is rasterised into a top-down map, rooms are recovered from it, and each object takes the room holding most of its extent (Fig.~\ref{fig:teaser}). The difficulty lies in the map itself. Walls are recorded only where the camera looked, so a gap in the boundary may be a doorway or a stretch of wall that was never observed; nothing distinguishes the two, and prior methods treat both as passages, merging rooms that should remain separate. We observe that both require the same treatment: a room should not extend across either, so both are closed and need not be distinguished. Both are also narrow relative to the rooms they separate, which we exploit twice. Such an opening closes under a small amount of boundary growth, and few sightlines cross it, so points in different rooms rarely see one another. We use the first to recover rooms and the second to assign objects to them. Rooms are obtained by Progressively thickening the boundary inward and freezing each free-space region once it becomes enclosed, so every opening seals at its own scale rather than at a radius fixed in advance; seeds are nothing but the camera poses, which removes the sampling heuristic and makes the segmentation deterministic. Detected door frames are treated as boundary wherever visibility is measured. Over 10 floors of 6 HM3D-Semantics scenes, scored against HOV-SG on identical top-down maps, we recover 74 rooms for 72 annotated regions (HOV-SG: 44), raising room $F_1$ from 0.741 to 0.890 at IoU $0.25$ at some cost in precision, and object-to-room ARI from 0.488 to 0.696 ($p=0.002$, ahead on every floor). We further demonstrate the pipeline on hand-held monocular video.
\end{abstract}
\begin{proteaser}
\centering
\resizebox{\textwidth}{!}{%
\begin{tikzpicture}[
  x=1cm, y=1cm, every node/.style={font=\scriptsize},
  grp/.append style={inner sep=5pt},
  grpours/.append style={inner sep=5pt},
]
\def\hs{1.25}
\def\hb{1.55}
\def\gs{0.40}
\def\gb{0.42}
\node[thumb, anchor=north west] (kf) at (0,0) {\includegraphics[height=\hs cm]{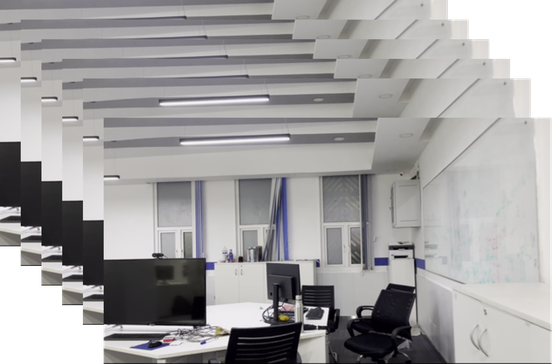}};
\node[thumb, anchor=north west] (pc) at ([xshift=\gs cm]kf.north east) {\includegraphics[height=\hs cm]{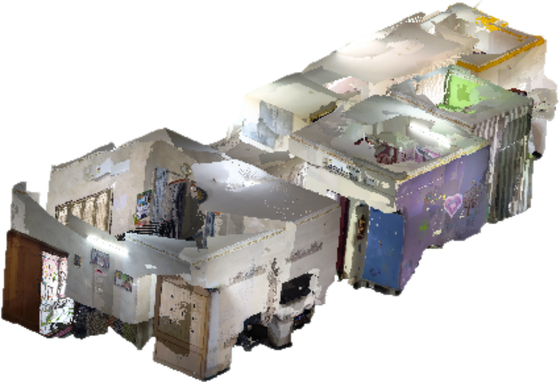}};
\node[thumb, anchor=north west] (wc) at ([xshift=\gs cm]pc.north east) {\includegraphics[height=\hs cm]{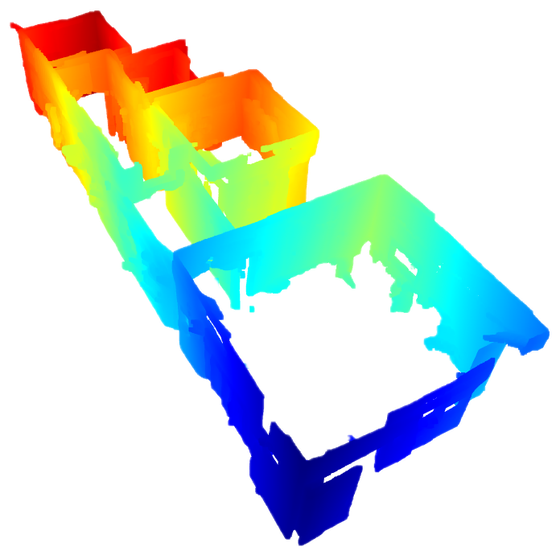}};
\node[sublab, anchor=north, align=center, inner sep=1pt] (kfl) at ([yshift=-1pt]kf.south) {monocular\\RGB video};
\node[sublab, anchor=north, align=center, inner sep=1pt] (pcl) at ([yshift=-1pt]pc.south) {dense point\\cloud};
\node[sublab, anchor=north, align=center, inner sep=1pt] (wcl) at ([yshift=-1pt]wc.south) {walls $+$\\windows};
\draw[flow] ([xshift=1pt]kf.east) -- ([xshift=-1pt]pc.west);
\draw[flow] ([xshift=1pt]pc.east) -- ([xshift=-1pt]wc.west);
\node[grplab, anchor=south west, inner sep=0pt] (frontlab) at ([yshift=4pt]kf.north west) {off-the-shelf front end: MASt3R-SLAM $+$ SAM3};
\begin{scope}[on background layer]
  \node[grp, fit=(frontlab)(kf)(kfl)(pc)(pcl)(wc)(wcl)] (front) {};
\end{scope}
\node[thumb, anchor=north west] (qa) at ([xshift=0.80cm]front.east|-kf.north) {\includegraphics[height=\hb cm]{clusters.png}};
\node[thumb, anchor=north west] (qb) at ([xshift=\gb cm]qa.north east) {\includegraphics[height=\hb cm]{iter_001.png}};
\node[thumb, anchor=north west] (qc) at ([xshift=\gb cm]qb.north east) {\includegraphics[height=\hb cm]{iter_009.png}};
\node[thumb, anchor=north west] (qd) at ([xshift=\gb cm]qc.north east) {\includegraphics[height=\hb cm]{iter_017.png}};
\node[thumb, anchor=north west] (qe) at ([xshift=\gb cm]qd.north east) {\includegraphics[height=\hb cm]{rays.png}};
\node[thumb, anchor=north west] (oa) at ([xshift=1.25cm]qe.north east) {\includegraphics[height=\hb cm]{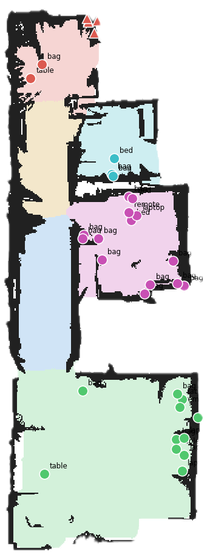}};
\foreach \n/\lab in {qa/{initial}, qb/{$t=1$}, qc/{$t=9$}, qd/{$t=17$}, qe/{sightlines}, oa/{rooms $+$ objects}}{
  \node[sublab, anchor=north, inner sep=1pt] (\n l) at ([yshift=-1pt]\n.south) {\lab};}
\foreach \f/\t in {qa/qb, qb/qc, qc/qd, qd/qe, qe/oa}{ \draw[flow] ([xshift=1.5pt]\f.east) -- ([xshift=-1.5pt]\t.west);}
\node[font=\tiny, text=black!70, align=center, anchor=south, inner sep=1pt] at ($(qe.east)!0.5!(oa.west)+(0,0.04)$) {$+$ object\\extents};
\node[grplabours, anchor=south west, inner sep=0pt] (bandlab) at ([yshift=4pt]qa.north west) {progressive boundary closure (ours)};
\begin{scope}[on background layer]
  \node[grpours, fit=(bandlab)(qa)(qal)(qb)(qbl)(qc)(qcl)(qd)(qdl)(qe)(qel)(oa)(oal)] (band) {};
\end{scope}
\draw[flow] ([xshift=2pt]front.east|-kf) -- ([xshift=-2pt]band.west|-kf);
\begin{scope}[on background layer]
  \node[draw=black!45, dotted, line width=.7pt, rounded corners=4pt, inner sep=6pt,
        fit=(front)(band)] (outer) {};
\end{scope}
\end{tikzpicture}
}
\caption{ProClosure recovers a room layer, and the objects inside it, from
uncalibrated monocular RGB video, with no depth, poses or calibration. Seeds
(dots) are the recorded camera positions. The bird's-eye boundary is thickened
until a pocket can admit no further seed, when it is declared a room and
frozen; each seed then casts sightlines that label free space, and each object
takes the room holding most of its reconstructed extent. 
}
\label{fig:teaser}
\end{proteaser}

\section{Introduction}
\label{sec:intro}

Hierarchical 3D scene graphs organise an environment into nested layers: object instances grouped into rooms, rooms into storeys. This structure makes them usable for navigation, planning and language-based interaction~\cite{armeni2019},~\cite{hughes2022hydra},~\cite{werby2024hovsg}. The room layer is pivotal within that hierarchy, since it is the level at which an agent acts on an instruction that names a room, such as navigating to the kitchen or searching it for a particular object. An error in the room partition propagates to every object grouping and every query posed above it.

Existing room layer methods obtain that partition from a top-down occupancy map and split it at architectural restrictions, by cutting a free-space graph at narrow passages \cite{hughes2022hydra}, by clearance-threshold flooding \cite{werby2024hovsg}, or by occupancy decomposition \cite{occusg2026}. Each assumes the boundary of the map is essentially complete, which holds for their input: posed range sensing recovers walls densely and continuously, so a gap in the boundary reliably marks a genuine opening such as a doorway.

The assumption fails for a reconstruction from monocular RGB video with no depth, poses, calibration or floor plan. A wall is recorded only where the camera observed it, so the boundary is broken in places that reflect the trajectory rather than the architecture. A break may be a doorway or a stretch of wall the camera never framed, and the two are geometrically indistinguishable. A method that partitions free space must therefore decide which breaks are architectural using evidence that does not contain the answer, and it merges the spaces behind those it misjudges. Prior monocular scene-graph work sidesteps this by adding inertial input \cite{monohydra} or a learned room-transition prior \cite{lexisg2026}. No prior room-layer method treats the incompleteness of the boundary as the central difficulty, which is present in any walk-based reconstruction, including the posed RGB-D walks on which we report quantitative results.

We take the ambiguity as given and resolve it with a single principle: treat every break by its width rather than its cause, and never classify an opening as a doorway. A doorway and an unframed gap are typically both narrow relative to the spaces they join, and such an opening is both easy to seal and hard to see through. We use each property in turn. To recover rooms we thicken the mapped walls inward and declare a room the moment a free-space region becomes enclosed, protecting it thereafter, so that a room is an event in a closure sequence rather than a region of one map. Each enclosed region is marked by a seed point. Rather than sample seeds from free space with a heuristic such as farthest-point sampling, we read them from the camera trajectory, since every recorded pose is a location the camera physically occupied. To place objects we use narrowness a second way. A wall with a gap in it still blocks the view between the two rooms it divides, even though the gap can be walked through, so the rooms stay distinguishable by what a point can see rather than by where it can go.

Closure alone leaves three characteristic errors: a region bounded by the raster edge rather than by a wall, a seed that relocates out of the region it witnesses, and a doorway the raster leaves open. Sec.~\ref{sec:cleanup} corrects each, the last using detected door frames as an additional input.

Our contributions are as follows:
\begin{enumerate}[label=\arabic*)]
  \item A formulation of room detection as \emph{progressive boundary closure}
with per-region freezing, in which a free-space component holding a seed
becomes a room at the step where it can no longer admit another, the seeds
being recorded camera positions rather than samples of the map. The room count
therefore emerges from the data, doorways and unobserved gaps are sealed
identically by width, and the segmentation is deterministic
(Sec.~\ref{sec:rooms}).
  \item An analysis of object-to-room assignment under an incomplete boundary, comparing a visibility-agreement rule against a surface-majority rule, evaluated directly as a clustering-agreement problem rather than indirectly through downstream retrieval (Secs.~\ref{sec:assign},~\ref{sec:setup}).
\end{enumerate}

\section{Related Work}

\subsection{Hierarchical 3D Scene Graphs} Scene graphs organise a reconstruction into entities and the relations between them. Armeni et al.~\cite{armeni2019} introduced the layered form now standard for buildings, in which a structure contains storeys, storeys contain rooms, and rooms contain objects. Kimera~\cite{rosinol2021kimera} and Hydra~\cite{hughes2022hydra} build such hierarchies online during exploration, deriving the room layer from a graph of free-space locations maintained as the environment is traversed. Later systems replace closed label sets with features from vision--language models: ConceptGraphs~\cite{gu2024conceptgraphs} builds open-vocabulary object maps without a room layer, HOV-SG~\cite{werby2024hovsg} recovers storeys, rooms and objects together and describes each with pooled visual features, and Clio~\cite{maggio2024clio} selects the granularity of the hierarchy according to a stated task. Each consumes depth measurements, inertial data, externally supplied poses, or some combination of the three, and the room layer in particular is built on geometry dense and complete enough that free space can be trusted. Work relaxing these constraints is recent and sparse. Mono-Hydra~\cite{monohydra} operates on monocular video with inertial support, and LEXI-SG~\cite{lexisg2026} constructs open-vocabulary hierarchies from RGB alone, deferring reconstruction until a room has been surveyed and identifying transitions between rooms from learned image features. Our concern is narrower than either: the room layer specifically, recovered from a reconstruction whose observed geometry is known to be incomplete. Our method takes as input a point cloud together with the camera trajectory that produced it, both supplied by the same reconstruction front end.

\subsection{Partitioning Indoor Space into Rooms} Clearance-based, connectivity-based and learned methods differ in the quantity they measure but agree in what they take a break in the boundary to mean; \suppref{B}{app:rwrooms} sets out each family and its citations. Each was developed for range sensing, where an aperture in the boundary is an aperture in the building. Under monocular reconstruction that correspondence does not hold: a wall the camera never observed leaves an opening of the same geometric character as a doorway, and where a surface is missed entirely, no boundary at all. The comparative study of the classical families~\cite{bormann2016} evaluates them on maps assumed complete, and supplies the overlap criterion we adopt in Sec.~\ref{sec:metrics}.

A separate line of work detects doorways explicitly, from geometry or from learned appearance, and cuts the map at what it finds. We use detected door frames only where visibility is measured, never as the primary partition criterion, so the closure still seals the unobserved gaps that no doorway detector will report.

\subsection{Morphological Segmentation and Scale} Thickening a boundary until an aperture seals, and recovering regions as connected components of what remains, are elementary morphological operations~\cite{serra1982,soille2003}; examining a structure across a graded sequence of scales rather than a single one is equally classical~\cite{matheron1975}. The difference lies in what is retained from the sequence. A single closing imposes one radius everywhere, trading the sealing of wide openings against the survival of small rooms; a granulometry sweeps radii but reports a single global distribution; the watershed transform~\cite{beucher1993watershed}, the usual route to morphological segmentation, resolves the entire domain at once from a fixed marker set. We instead extract each region at the scale at which its own boundary closes, exempt it from further growth thereafter, and carry witnesses through the contraction so that regions remain detectable as the free space around them shrinks.

\section{Methodology}

\begin{figure*}[!t]
\centering
\begin{adjustbox}{max width=\textwidth}
\begin{tikzpicture}[
  x=1cm, y=1cm,
  every node/.style={font=\footnotesize},
  proc/.append style={minimum height=5.8mm, inner xsep=5pt},
  data/.append style={inner xsep=3pt},
  term/.append style={inner xsep=3.5pt},
  rowlab/.style={font=\scriptsize\bfseries, text=black!55, anchor=east},
]
\node[thumb, anchor=west] (kfi)  at (0.95,0)   {\includegraphics[height=1.40cm]{input_frames.png}};
\node[proc,  anchor=west] (slam) at (3.40,0)   {MASt3R-SLAM~\cite{mast3rslam}};
\node[thumb, anchor=west] (pci)  at (6.70,0)   {\includegraphics[height=1.84cm]{mast3r_slam.png}};
\node[proc,  anchor=west] (sam)  at (9.85,0)   {SAM3~\cite{sam3}};
\node[proc,  anchor=west] (lift) at (12.15,0)  {lift to 3D};
\node[thumb, anchor=west] (wci)  at (14.80,0)  {\includegraphics[height=1.84cm]{wall_cloud.png}};

\node[rowlab] at (0.80,0) {(a)};
\node[sublab, anchor=north, inner sep=1pt] at ([yshift=-2pt]kfi.south) {input video};
\node[sublab, anchor=north, inner sep=1pt] at ([yshift=-2pt]pci.south) {dense cloud};
\node[sublab, anchor=north, inner sep=1pt] at ([yshift=-2pt]wci.south) {$\mathcal{P}_w$ wall\,/\,window};

\draw[dflow] (kfi.east)  -- (slam.west);
\draw[dflow] (slam.east) -- (pci.west);
\draw[dflow] (pci.east)  -- (sam.west);
\draw[dflow] (sam.east)  -- (lift.west);
\draw[dflow] (lift.east) -- (wci.west);

\node[data, anchor=west] (traj) at (0.95,-2.10) {camera trajectory};
\node[data, anchor=west] (dlt)  at (4.15,-2.10) {door frames $\Delta$};
\node[data, anchor=west] (opc)  at (7.35,-2.10) {object clouds};
\node[proc, anchor=west] (trk)  at (9.85,-2.10) {voxel-IoS tracker};
\node[term, anchor=west] (trks) at (13.10,-2.10) {tracks $\{o_j\}$, extents $\{U_j\}$};
\node[rowlab] at (0.80,-2.10) {(b)};

\draw[dflow] (slam.south) -- ++(0,-1.28) -| ([xshift=8mm]traj.north);
\draw[dflow] (lift.south) -- ++(0,-1.28) -| ([xshift=8mm]dlt.north);
\draw[dflow] (lift.south) -- ++(0,-1.28) -| ([xshift=8mm]opc.north);
\draw[dflow] (opc.east) -- (trk.west);
\draw[dflow] (trk.east) -- (trks.west);

\node[data,  anchor=west] (pwin) at (0.95,-3.62) {$\mathcal{P}_w$};
\node[proc,  anchor=west] (grv)  at (2.45,-3.62) {gravity $+$ Manhattan};
\node[proc,  anchor=west] (prj)  at (6.30,-3.62) {height cut $+$ BEV};
\node[thumb, anchor=west] (bevi) at (9.85,-3.62) {\includegraphics[height=1.66cm]{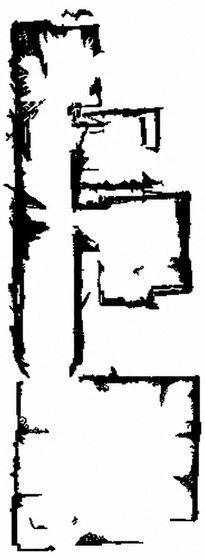}};
\node[sublab, anchor=west, inner sep=3pt] at (bevi.east) {top-down map $(W,F)$};
\node[term,  anchor=west] (outs) at (13.10,-3.62) {$(W,F)\;\cup\;\Delta$, seeds, $\{U_j\}$};
\node[rowlab] at (0.80,-3.62) {(c)};

\draw[dflow] (pwin.east) -- (grv.west);
\draw[dflow] (grv.east)  -- (prj.west);
\draw[dflow] (prj.east)  -- (bevi.west);

\begin{scope}[on background layer]
  \node[draw=blue!45!black, dashed, dash pattern=on 2.4pt off 1.8pt, rounded corners=4pt,
        fill=blue!3, inner sep=6pt,
        fit=(kfi)(slam)(pci)(sam)(lift)(wci)(traj)(dlt)(opc)(trk)(trks)
            (pwin)(grv)(prj)(bevi)(outs)] (box) {};
\end{scope}
\node[font=\scriptsize\scshape, text=blue!45!black, anchor=north west, inner sep=2pt]
      at ([shift={(2pt,-1pt)}]box.north west) {};

\end{tikzpicture}
\end{adjustbox}

\caption{Reconstruction and projection (Sec.~\ref{sec:bev},
\supprefp{B}{app:overview}), built from established components. (a)
MASt3R-SLAM~\cite{mast3rslam} returns keyframes, a dense point cloud and the
camera trajectory. (b) SAM3~\cite{sam3} segments each keyframe against a fixed
eleven-label prompt set and the masks are lifted through the per-pixel 3D points
the front end supplies, so that \emph{wall} and \emph{window} accumulate into the
structural cloud $\mathcal{P}_w$, \emph{door frame} gives the door frames $\Delta$
the visibility tests seal on (Sec.~\ref{sec:cleanup}), and the remaining prompts
become object observations tracked by voxel-IoS matching
(\supprefp{A}{app:recon}). (c) $\mathcal{P}_w$ is aligned, height-cut and
rasterised into the binary top-down map $(W, F)$. 
}

\label{fig:pipeline-data}
\end{figure*}

We construct a room-level scene graph of an indoor environment from a single monocular RGB video, assuming no depth, externally supplied poses, calibration or floor plan. Such a reconstruction is fixed only up to a global similarity, so no absolute length is available, and every spatial threshold below is expressed relative to a quantity measured from the scene itself.

The method has three stages. (i)~\emph{Reconstruction and segmentation}: a
feed-forward pointmap SLAM front end returns the keyframe trajectory
$\mathrm{T}$ and a dense cloud $\mathcal{P}$. A promptable open-vocabulary
model separates structural surfaces from objects, both lifted through the
per-pixel 3D points the front end supplies. This yields the structural cloud
$\mathcal{P}_w$, the detected door frames, and the object tracks $\mathcal{O}$.
(ii)~\emph{Canonical projection} (Sec.~\ref{sec:bev}): $\mathcal{P}_w$ is placed
upright, axis-aligned and rasterised onto a square grid $\Omega$, giving a
binary top-down map with occupied set $W \subset \Omega$ and free space
$F = \Omega \setminus W$. The full cloud $\mathcal{P}$ is rasterised on the same grid by the same
$\pi_Q$, and the outer contour of that projection filled, giving the building
footprint $\Omega_{\mathrm{fp}} \subset \Omega$, which records the extent the
capture covers. (iii)~\emph{Rooms and objects} (Secs.~\ref{sec:rooms}--\ref{sec:assign}): rooms
are recovered from $W$ by the proposed progressive boundary closure, seeded by
the projected poses. Each object is assigned to the room holding most of its
reconstructed extent.


\subsection{Canonical Alignment and Top-Down Projection}
\label{sec:bev}
Figure~\ref{fig:pipeline-data} traces one capture from keyframes~(a) through lifted masks~(b) to the binary map~(c) this section produces. Given a point
cloud from monocular video, the map must be formed along the true vertical,
since projecting along any other axis collapses walls onto the floor. We
estimate the canonical frame from $\mathcal{P}$, apply it to $\mathcal{P}_w$,
and rasterise.

\paragraph{Canonical frame and height cut.}
We estimate a rotation $Q$ that puts the floor normal on the vertical axis and
aligns the dominant wall direction with the grid, then cut the transformed cloud
at the mid-height of its own vertical extent. The plane fitting, the orientation
histogram and the height argument are given in \suppref{A}{app:canon}.

\paragraph{Rasterisation.}
We write $\pi_Q(x)$ for the image of a 3D point $x$ under $Q$ followed by
orthographic projection onto the horizontal plane; $\pi_Q$ is the single
transform that carries the reconstruction onto the map, and every later
projection uses it. The structural cloud is transformed by $Q$, projected by
$\pi_Q$, enclosed in an axis-aligned square with fixed relative padding, and
discretised into the grid $\Omega$ of side $N$ cells. A cell $u \in \Omega$ is
occupied when at least one projected point falls in it, giving a binary rather
than a density map:
\begin{equation}
  W = \bigl\{\, u \in \Omega : \exists\, x \in \mathcal{P}_w,\ \pi_Q(x) \in u
  \,\bigr\}, \qquad F = \Omega \setminus W .
  \label{eq:occupancy}
\end{equation}
Because the extent is measured from the scene and $N$ is fixed, one cell
subtends a constant fraction of the scene, so any threshold expressed in cells
is invariant to the unknown scale of $\mathcal{P}$; every parameter in
Secs.~\ref{sec:rooms}--\ref{sec:assign} is accordingly a cell count or a
dimensionless ratio. Isolated occupied cells from reconstruction noise are then
removed by a connected-component area filter, which preserves extended wall
lines.

\paragraph{Footprint.}
The map records walls alone, so the blank margin outside the building is free
space in it exactly as a room is, and nothing in the raster separates the two.
To recover the distinction we project the full cloud $\mathcal{P}$, rather than
the structural cloud $\mathcal{P}_w$, by the same $\pi_Q$ onto the same grid and
under the same height cut, then smooth that raster, close it, and fill its
outermost contour. The floors and furniture $\mathcal{P}_w$ omits are what mark
the interior. The result is the building footprint
$\Omega_{\mathrm{fp}} \subset \Omega$, the area the reconstruction covers.
Sec.~\ref{sec:cleanup} uses it to separate a region enclosed by walls from one
that merely ran out of raster, and Sec.~\ref{sec:delineate} uses it to confine
the interior fill.

\paragraph{Door frames.}
The door frames of stage~(i) are carried onto the grid by the same $\pi_Q$.
Each detected frame is reduced to the filled convex hull of its projected
cells, so that an opening is sealed rather than merely dotted; the door panel
is not used, only the frame around it. An instance is discarded if fewer than
half its cells fall on $W$, which removes detections not attached to a wall, or
if its hull exceeds a fixed area, which removes over-large ones. The union of
the surviving hulls is the cell set $\Delta$ that Sec.~\ref{sec:cleanup} uses.
It is kept apart from $W$, since it is boundary only where visibility is
measured and never during the closure.

\subsection{Room Detection by Progressive Boundary Closure}
\label{sec:rooms}
We define a room not as a region of one map but as an event in a sequence. The boundary is thickened one cell at a time, and at each step the free space splits into connected components. A component is \emph{saturated} when it is packed to its own clearance limit, holding no cell that is both far enough from the boundary and far enough from the seeds already inside it to admit another well-separated seed. A component holding a seed is declared a room at the first step at which it saturates, and is frozen thereafter, so subsequent thickening does not enter it. The clearance and separation thresholds are given in Sec.~\ref{sec:closure}.

Because a component is declared when its own boundary closes, the thickness at which that happens is a property of the region rather than a value fixed in advance for the whole floor. This is what separates the method from a single morphological closing, the natural baseline. A radius large enough to seal a wide archway also consumes a small room, while one that preserves small rooms leaves wide apertures open. Real floors mix aperture widths, so no single radius is correct, and freezing each region as it encloses gives each one its own effective radius.

\subsubsection{Progressive Closure}
\label{sec:closure}

The boundary grows in small steps, indexed by $t$. At every step all walls are
thickened by a disc $B_\rho$ of radius $\rho$, so the free space beside each
wall shrinks. Rooms already declared are exempt, and growth stops at their edge.
Let $\Phi^{(t)}$ hold the cells of every room declared up to step $t$ by the
rule below, with $\Phi^{(0)} = \emptyset$ and $W^{(0)} = W$; regions enter
$\Phi$ whole and never leave. Each step dilates $W^{(t-1)}$ by $B_\rho$
everywhere outside $\Phi^{(t-1)}$ and leaves the cells inside it untouched
(Eq.~(10), Appendix~M), so a room keeps the extent it had when it sealed.
Thickening can leave a seed too close to a wall; it is moved back to a cell with
space for it (Sec.~\ref{sec:seeds}) before the test below is applied.

\paragraph{When a region becomes a room.}
A pocket, meaning a connected component of the free space outside $\Phi$, is declared a room when it can no longer take another seed. The rule that places seeds is what makes this meaningful. A seed must lie at least $\kappa$ cells from the nearest wall and at least $\sigma$ cells from every other seed, so each seed needs a patch of open floor to itself; $\kappa$ is the clearance and $\sigma$ the separation. How many seeds a pocket can take is therefore a measure of how much open floor it has. As the walls thicken the pocket shrinks and fewer cells can host a seed. When none can, the pocket has closed down to the floor its seeds already occupy. We call it a room at that step.

The qualifying cells $V^{(t)}$ are those lying at least $\kappa$ from the boundary, measured by the distance transform of Eq.~\eqref{eq:edt}, and at least $\sigma$ from every current seed (Eq.~\eqref{eq:admissible}). A pocket is saturated when it holds none of them. Every saturated pocket containing a seed is declared a room and added to $\Phi$ (Eq.~\eqref{eq:freeze}), taken over $\mathcal{C}^{(t)}$, the connected components of $F^{(t)} \setminus \Phi^{(t-1)}$. Growth stops when all seeds lie inside declared rooms, when dilation no longer changes $W$, or at a cap $t_{\max}$.

\paragraph{Why this separates rooms.}
A doorway is the narrowest part of the boundary between two rooms, so the growing walls meet there first. When it closes, the pocket spanning both rooms splits in two and each half saturates on its own. A wide opening survives longer than a narrow one, so every opening seals at the scale set by its own width. Two situations defeat this. If an opening is wider than the shallower room is deep, that room fills and is declared before the opening closes. If two spaces meet with no constriction at all, there is no width minimum for the growth to find and they never separate. Sec.~\ref{sec:res-fail} reports both.

\subsubsection{Seeds as Region Witnesses}
\label{sec:seeds}

\begin{algorithm}[!t]
\caption{Room detection by progressive boundary closure}
\label{alg:rooms}
\begin{algorithmic}[1]
\Require occupied set $W$; door frames $\Delta$; footprint $\Omega_{\mathrm{fp}}$; trajectory $\mathrm{T}$; parameters $\rho, \kappa, \sigma, \mu, t_{\max}, k, n_{\min}, \phi, \tau$
\Ensure  rooms $\{\mathcal{R}_i\}$ with seed clusters $\{\mathcal{S}_i\}$
\State $\mathcal{S} \gets \textsc{TrajectorySeeds}(\mathrm{T}, W, \sigma, \kappa)$ \Comment{Sec.~\ref{sec:seeds}}
\State $\Phi \gets \emptyset$;\quad $W^{(0)} \gets W$
\For{$t = 1$ \textbf{to} $t_{\max}$}
  \State $W^{(t)} \gets$ dilate $W^{(t-1)}$ by $B_\rho$, protecting $\Phi$ \Comment{Eq.~\eqref{eq:dilate}}
  \State $D^{(t)} \gets$ EDT of $\Omega \setminus W^{(t)}$ \Comment{Eq.~\eqref{eq:edt}}
  \ForAll{undeclared seeds $p$ with $D^{(t)}(p) < \kappa$ (to a fixed point)}
    \State $p \gets \arg\min_{q \in A^{(t)}(p)} d_g(p,q)$ if $A^{(t)}(p) \neq \emptyset$ \Comment{Eq.~\eqref{eq:relocate}}
  \EndFor
  \State $V^{(t)} \gets \{\, u \in F^{(t)} \setminus \Phi : D^{(t)}(u) \ge \kappa,\ d_{\mathcal{S}}^{(t)}(u) \ge \sigma \,\}$ \Comment{Eq.~\eqref{eq:admissible}}
  \ForAll{components $C$ of $F^{(t)} \setminus \Phi$ with $C \cap \mathcal{S}^{(t)} \neq \emptyset$}
    \If{$C \cap V^{(t)} = \emptyset$}
      \State $\Phi \gets \Phi \cup C$ \Comment{saturated; Eq.~\eqref{eq:freeze}}
    \EndIf
  \EndFor
  \State \textbf{if} all seeds lie in $\Phi$ \textbf{or} $W^{(t)} = W^{(t-1)}$
         \textbf{then break}
\EndFor
\State assign each seed to the component of $\Phi$ containing it, giving clusters $\{\mathcal{S}_i\}$
\State score every seed from its sampled position $p^{\mathrm{init}}$
\Statex \hrulefill\ \emph{corrections} (Sec.~\ref{sec:cleanup})\ \hrulefill
\ForAll{clusters $i$}
  \If{$|\mathcal{R}_i \setminus \Omega_{\mathrm{fp}}| / |\mathcal{R}_i| > \phi$}
    \State discard region $i$; return its seeds unassigned \Comment{Eq.~\eqref{eq:unsealed}}
  \EndIf
\EndFor
\State $c_i \gets$ centroid of region $i$;\quad evaluate $a(\cdot,\cdot)$ on $W \cup \Delta$
\ForAll{seeds $p$ in ascending order of $a(p, c_{i(p)})$, unassigned seeds first} \Comment{$i(p)$: index of $p$'s current region, so $c_{i(p)}$ is its centroid}
  \State \textbf{if} $p$ lies on $\Delta$ \textbf{then} discard $p$; \textbf{continue}
  \If{$a(p, c_{i(p)}) \ge \tau$}
    \State \textbf{continue} \Comment{sees its own region}
  \EndIf
  \State $i^{\star} \gets \arg\max_i a(p, c_i)$
  \State \textbf{if} $a(p, c_{i^{\star}}) \ge \tau$ \textbf{then} move $p$ to cluster $i^{\star}$ \textbf{else} start a new cluster at $p$
\EndFor
\ForAll{clusters with $\ge n_{\min}$ seeds}
  \State discard seeds with displacement $> \bar{d} + k\,s_d$ \Comment{outlier removal}
\EndFor
\Statex \hrulefill
\State $\{\mathcal{R}_i\} \gets \textsc{Delineate}(\{\mathcal{S}_i\}, W \cup \Delta)$ \Comment{Sec.~\ref{sec:delineate}}
\State \Return $\{\mathcal{R}_i\}, \{\mathcal{S}_i\}$
\end{algorithmic}
\end{algorithm}

Equation~\eqref{eq:freeze} declares a component only when it contains a seed, so a seed \emph{witnesses} that a bounded region is present. Seeds do not decide where the regions are; Eq.~\eqref{eq:freeze} does. What a seed must do is stay inside its region while the free space contracts, because a region that loses its last witness is never declared.

\paragraph{Placement.}
The seeds are the recorded camera positions, carried onto the grid by the same projection $\pi_Q$ that formed the map. Poses lying in this floor's height band are visited in capture order, and a pose becomes a seed if it has clearance at least $\kappa$ from the boundary and separation at least $\sigma$ from every seed already kept. Clearance is measured by the Euclidean distance transform of free space, which gives for each cell $p$ its distance to the nearest occupied cell $q$,
\begin{equation}
  D^{(t)}(p) = \min_{q \in W^{(t)}} \lVert p - q \rVert_2 ,
  \label{eq:edt}
\end{equation}
so a seed starts well inside its region rather than against a wall. No eligibility pool is built and no interior test is applied, because a recorded pose is a place the camera physically occupied and is therefore free space by construction. The procedure makes no random choice, so the segmentation is deterministic and there is no seed to report variance over. Sampling the map instead, either uniformly or by farthest point sampling, leaves Eqs.~\eqref{eq:dilate}--\eqref{eq:freeze} untouched and is compared in Sec.~\ref{sec:res-ablation}.

\paragraph{Relocation.}
As the boundary thickens it encroaches on seeds; an undeclared seed whose clearance falls below $\kappa$ moves to the geodesically nearest admissible cell, measured within free space and excluding declared regions so it cannot cross the boundary or enter a resolved room. The admissible set and its displacement cap are given in \suppref{C}{app:relocate}.

\subsubsection{Rooms and Cleanup}
\label{sec:cleanup}

On termination the components of $\Phi$ are the rooms $\{\mathcal{R}_i\}$, and each seed joins the component containing it, giving clusters $\{\mathcal{S}_i\}$. Three corrections are then applied, all given in full in \suppref{D}{app:cleanup}. \emph{Dropping unsealed clusters}: a pocket bounded by the edge of the raster rather than by a wall never closed, so it is not a room; it is discarded and its seeds are set free. \emph{Reassigning seeds to centroids}: a seed that cannot see the region it sits in, whether it was set free above or drifted while relocating, joins the region it can see, and founds a new one if it can see none. \emph{Sealing doorways}: detected door frames count as boundary in every visibility test, so a sightline cannot pass through an open doorway. Seeds that travelled much further than their room-mates are discarded alongside.

\paragraph{Doorways as boundary.}
Every visibility test is computed on $W \cup \Delta$ rather than on $W$: both the re-homing above and the sightlines of Sec.~\ref{sec:delineate}. Through an open doorway a fan reaches into the next room, so two seeds on opposite sides score high agreement only because the door is open. The closure itself is untouched and still grows the raster it was given. Sec.~\ref{sec:res-ablation} reports what this contributes.

\subsubsection{Room Delineation}
\label{sec:delineate}

Freezing settles how many rooms there are and which seeds belong to each, but not where their boundaries lie. We recover the boundaries by growing each room outward from its own seeds until the free space is covered.

Each seed casts the sightlines of Sec.~\ref{sec:assign:vis} on $W \cup \Delta$. Where rays from two seeds of different rooms cross, the ray from the farther seed is removed, so a sightline reaching through an opening is discounted at its source rather than where it lands. Every surviving sightline votes its seed's room along its length, and each free cell takes the room with the most votes through it.

Cells that no sightline reaches are filled by propagating room identity through free space to the geodesically nearest labelled cell, so a room grows within the boundary and never leaks across a wall. The fill is confined to $\Omega_{\mathrm{fp}}$ eroded by a few cells, which stops a room touching an outer wall from flooding the exterior of a walls-only map; the erosion is needed because the footprint is a filled contour whose edge sits slightly outside the true wall.

The result is one room label per free cell, and hence one per reconstructed 3D point. The pass distributes area among rooms already decided by Eq.~\eqref{eq:freeze}: it cannot create a room or merge two. It can remove one, in that a region holding too small a share of the labelled area is dropped. The count is otherwise settled by Eq.~\eqref{eq:freeze} and the corrections of Sec.~\ref{sec:cleanup}, which may discard a region or found one.

The procedure is Algorithm~\ref{alg:rooms}; three properties that follow from it are given in Appendix~\ref{app:properties}.

\subsection{Object Placement}
\label{sec:assign}

\begin{figure}[!t]
\centering

\begin{adjustbox}{max width=\textwidth}
    
\begin{tikzpicture}[
  every node/.style={font=\scriptsize},
  proc/.append style={minimum height=3.6mm, inner xsep=3pt, inner ysep=2pt},
  ours/.append style={minimum height=3.6mm, inner xsep=3pt, inner ysep=2pt},
  data/.append style={inner xsep=3pt, inner ysep=2pt},
  term/.append style={inner xsep=3pt, inner ysep=2pt},
  grpours/.append style={inner sep=7pt},
]
\node[data, inner xsep=5pt] (bev) {top-down map\\$(W,F)\;\cup\;\Delta$};
\node[ours, right=7mm of bev] (clo) {progressive closure};
\node[data, right=7mm of clo, inner xsep=5pt] (sds) {seed groups $\{\mathcal{S}_i\}$};
\draw[dflow] (bev) -- (clo);
\draw[dflow] (clo) -- (sds);

\node[ours, below=3mm of bev.south west, anchor=north west] (den) {ray vote $+$ BFS fill};
\node[term, right=9mm of den] (rms) {rooms $\{\mathcal{R}_i\}$};
\draw[dflow] (sds.south) -- ++(0,-1.4mm) -| (den.north);
\draw[dflow] (den) -- (rms);

\node[data, below=4mm of den.south west, anchor=north west, inner xsep=5pt] (ext) {extents $\{U_j\}$};
\node[ours, right=7mm of ext] (sm) {surface majority};
\draw[dflow] (ext) -- (sm);
\draw[dflow] (rms.south) -- ++(0,-1.8mm) -| (sm.north);

\node[proc, below=2mm of ext.south west, anchor=north west, draw=black!45, dashed,
      dash pattern=on 2pt off 1.5pt, inner xsep=6pt] (va)
      {visibility agreement (Sec.~\ref{sec:res-ablation})};
\node[term, right=5mm of sm.east, anchor=west] (out) {assignments $f$};
\draw[dflow] (sm) -- (out);

\node[term, right=5mm of sm.east, anchor=west] (out) {assignments $f$};
\draw[dflow] (sm) -- (out);

\coordinate (vaOut) at ([yshift= 1.2mm]va.east);   
\coordinate (vaIn)  at ([yshift=-1.2mm]va.east);   

\coordinate (varise) at ([xshift=-3mm]out.south);  
\coordinate (fblane) at ([xshift= 4mm]out.east);   

\draw[dflow, rounded corners=1.5pt]
      (vaOut) -- (vaOut -| varise) -- (varise);

\draw[dflow, rounded corners=1.5pt]
      (rms.east) -- (rms.east -| fblane)
      -- node[sublab, pos=0.6, right]{fallback} (vaIn -| fblane)
      -- (vaIn);
      
\begin{scope}[shift={($(ext.south west)+(0,-1.05)$)}, x=1cm, y=1cm]
\def\pw{1.0}\def\gx{0.30}
\pgfmathsetmacro{\ph}{\pw/0.371}
\foreach \f/\k/\lab [count=\i from 0] in {%
    clusters/a/{initial}, iter_001/b/{$t=1$},
    iter_009/c/{$t=9$}, iter_017/d/{$t=17$},
    rays/e/{sightlines}}{
  \node[thumb, anchor=north west] (p\k) at ({\i*(\pw+\gx)},0)
        {\includegraphics[width=\pw cm]{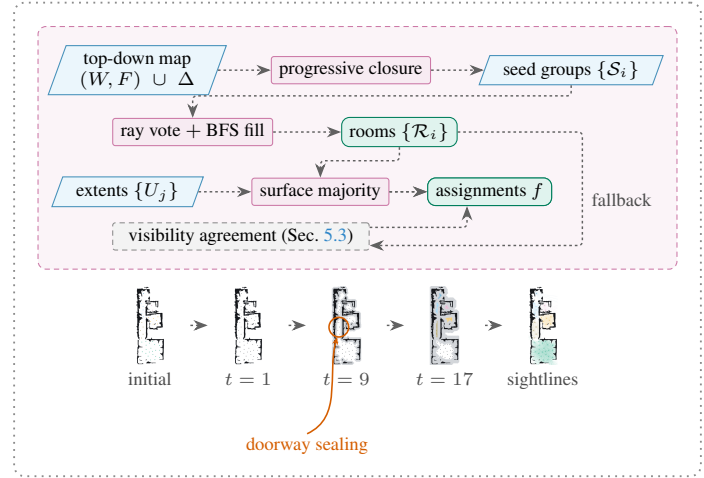}};
  \node[sublab, anchor=north, inner sep=1pt] at ([yshift=-2pt]p\k.south) {\lab};
}
\foreach \f/\t in {pa/pb, pb/pc, pc/pd}{
  \draw[dflow] ([xshift=1pt]\f.east) -- ([xshift=-1pt]\t.west);}
\draw[dflow] ([xshift=1pt]pd.east) -- ([xshift=-1pt]pe.west);


\coordinate (dw) at ([xshift=0.40*\pw cm, yshift=-0.200*\ph cm]pc.north west);
\draw[semithick, oiVermillion] (dw) circle (0.13);

\coordinate (dwtip) at ($(dw)+(-100:0.13)$);                       
\coordinate (dwlab) at ([xshift=-0.42cm, yshift=-0.92cm]dw |- pc.south);

\draw[semithick, oiVermillion, -{Stealth[length=3pt]}]
      (dwlab) to[out=25, in=-105] (dwtip);
\node[anchor=north, inner sep=1pt, text=oiVermillion, font=\scriptsize]
      at (dwlab) {doorway sealing};
      
\end{scope}

\begin{scope}[on background layer]
  \node[grpours, fit=(bev)(clo)(sds)(den)(rms)(ext)(sm)(va)(out)] (ours) {};
  \node[draw=black!45, dotted, line width=.7pt, rounded corners=4pt, inner sep=16pt,
        fit=(bev)(clo)(sds)(den)(rms)(ext)(sm)(va)(out)(pa)(pb)(pc)(pd)(pe)(dwlab)] (chainbox) {};
\end{scope}
\end{tikzpicture}%
\end{adjustbox}

\caption{Progressive closure and assignment. \emph{Top}: the boundaries of $W$ are grown until every free-space pocket saturates, seeded by the projected camera trajectory rather than by sampling the map (Sec.~\ref{sec:rooms}); the retained sightlines of the seed groups $\{\mathcal{S}_i\}$, cast on $W \cup \Delta$ so that a detected door frame seals a doorway the raster leaves open, vote for a label at each free cell and a breadth-first fill restricted to free space completes the dense rooms $\{\mathcal{R}_i\}$. Surface majority (Sec.~\ref{sec:surfmaj}) consumes those dense labels; the dashed visibility-agreement rule it is compared against (Sec.~\ref{sec:assign:vis}) does not, consulting them only through the fallback edge taken when no room reaches a nonzero agreement. \emph{Bottom}: the closure sequence on one floor, with one doorway circled as it seals.}
\label{fig:pipeline}
\end{figure}
Characterising a location by what can be seen from it originates in spatial analysis, where the isovist --- the region visible from a point~\cite{benedikt1979, davis1979} --- measures enclosure and openness, and visibility graphs extend the construction to relations between locations~\cite{turner2001}. We put mutual visibility to a narrower use: deciding which room an object belongs to.

That step has received little dedicated attention. Systems maintaining a room layer generally test whether an object's centroid falls within a region~\cite{werby2024hovsg, hughes2022hydra}, and assess the outcome only indirectly, through the success of queries naming both a room and an object~\cite{werby2024hovsg, keysg2025}. We are not aware of a standard metric for the assignment itself, and specify one in Sec.~\ref{sec:setup}.

Each track $o_j$ has a centroid $c_j$ projecting to $\pi(o_j) = \pi_Q(c_j)$, and a reconstructed extent projecting to a set of grid cells. The obvious way to place it is containment: take the room whose region holds $\pi(o_j)$. That test is brittle here. A centroid built from few or oblique views carries error comparable to the width of an opening, which puts an object near a boundary on the wrong side of it, and the delineated region is itself an estimate near that same boundary, so the test can fail on either input. We therefore use rules that do not depend on one point falling inside one region, and that degrade as the centroid drifts rather than switching at a region edge. Sec.~\ref{sec:surfmaj} gives the rule used throughout, and Sec.~\ref{sec:assign:vis} the alternative it is compared against in Sec.~\ref{sec:res-ablation}.

\subsubsection{Surface Majority}
\label{sec:surfmaj}

The rule used throughout replaces the object's centroid with its reconstructed extent. Each track carries the surface points accumulated over the views in which it was seen; projecting them onto the grid gives $U_j$, the cells the object covers from above, counted with multiplicity. The object takes the room holding the largest share of them,
\begin{equation}
  f(o_j) = \arg\max_{i} \, \bigl\lvert U_j \cap \mathcal{R}_i \bigr\rvert .
  \label{eq:surfmaj}
\end{equation}
Every projected point votes for the room label at its own cell, so the vote is read from the same two-dimensional labelling on which the rooms were delineated, while an object's weight follows its reconstructed surface rather than its floor area alone.

The support $\mathrm{supp}(o_j) = \max_i \lvert U_j \cap \mathcal{R}_i \rvert / \lvert U_j \rvert$ records how decisive the vote was: unity for an object lying wholly inside one room, near one half for one straddling a boundary. Points falling on the boundary carry no room label and are ignored. When every point falls on the boundary, which happens for objects mounted flush against a wall whose extent is thinner than the reconstructed wall, each projected cell takes the nearest room measured through free space, and the object takes the majority over them.

\subsubsection{Visibility agreement}
\label{sec:assign:vis}

From a cell $p$ we cast $K$ rays at angles $\theta_k = 2\pi k / K$, each stopping at the first cell of $W \cup \Delta$, at the edge of $\Omega$, or at a length cap $\ell$, set separately for seed and object fans. Writing $r_k^{o}(p)$ for the $k$th ray of the object fan at $p$ and $r_l^{s}(q)$ for the $l$th of the seed fan at $q$, agreement is the fraction of the object's rays that meet some ray of the seed's,
\begin{equation}
  a(p, q) = \frac{1}{K} \sum_{k=1}^{K}
  \mathbb{1}\!\left[\, \exists\, l : r_k^{o}(p) \cap r_l^{s}(q) \neq \emptyset \,\right]
  \in [0, 1].
  \label{eq:agreement}
\end{equation}
Two points inside one enclosed region see the same boundary, so their fans cross often and $a$ is high. Two points separated by an intact wall see disjoint stretches of boundary and never meet. Across a narrow opening only the few rays passing through it can cross, so $a$ stays small whatever the opening's cause, which is the property the rule relies on.

Each object is scored per room by mean agreement with that room's seeds,
\begin{equation}
  s_i(o_j) = \frac{1}{|\mathcal{S}_i|} \sum_{q \in \mathcal{S}_i}
  a\bigl(\pi(o_j), q\bigr),
  \label{eq:objscore}
\end{equation}
and is assigned to the highest-scoring room when that score reaches a threshold $\tau_a$, falling back to the room of the nearest free cell otherwise. Appendix~\ref{app:assign-alg} gives the remaining details, and Sec.~\ref{sec:res-ablation} reports the fallback rate and the comparison against surface majority.

\section{Experimental Setup}
\label{sec:setup}

\subsection{Datasets}
\label{sec:datasets}

We evaluate on two sources. Quantitative results use annotated scenes whose ground truth exists independently of us; hand-held captures are shown qualitatively, under the conditions the method is designed for.

\paragraph{Annotated scenes (quantitative).}
We use scenes from HM3D--Semantics v0.2 (val), the dataset adopted by HOV-SG~\cite{werby2024hovsg}, so that our room numbers can be read against theirs. For each scene we render an RGB-D trajectory with the simulator. So that both methods see identical maps, stage~(i) of the pipeline is replaced by the posed RGB-D fusion HOV-SG~\cite{werby2024hovsg}itself consumes: the structural cloud and the trajectory are taken from the rendered walk rather than estimated from RGB. Stages~(ii) and~(iii), the contributions, are unchanged, and the monocular front end is exercised separately in Sec.~\ref{sec:res-qual}. The set contains \textbf{6 scenes with 10 floors}, a subset of HOV-SG's eight, holding \textbf{2264} annotated object instances across \textbf{105} annotated regions. The door frames of Sec.~\ref{sec:cleanup} come from the same annotations, and Sec.~\ref{sec:baselines} states what this gives our method that the baseline does not receive.


\paragraph{Hand-held captures (qualitative).}
We record hand-held videos of real interiors with a mobile phone, covering 3 distinct layouts, used only for the qualitative results of Sec.~\ref{sec:res-qual} and the failure analysis.

\subsection{Evaluation Protocol and Metrics}
\label{sec:metrics}

\paragraph{Room partitioning.} Predicted and ground-truth rooms are sets of cells in the top-down grid. We report two families of scores, which differ in how a prediction is paired with a region and are not interchangeable.

\paragraph{Object-to-room assignment.} Scored over the 2264 annotated instances by Hungarian-matched accuracy, plus Adjusted Rand Index (ARI) and Normalized Mutual Information (NMI). Neither ARI nor NMI needs a correspondence between predicted rooms and annotated regions, so both stay comparable when the methods predict different numbers of rooms, while still penalising a mismatched count. Both methods are scored under the identical assignment rule, so the object columns of Table~\ref{tab:rooms-main} isolate the room segmentation; the assignment rule itself is isolated in Table~\ref{tab:ablation}.

\paragraph{Statistical treatment.} Per-floor differences are tested with a paired Wilcoxon signed-rank test. With ten paired samples the test cannot return $p < 0.002$, so we report the per-floor win count alongside it. Room and object scores can disagree, since a room with an imprecise boundary may still contain all of its objects, which is why both are given.

\subsection{Baselines and Comparison Methods} \label{sec:baselines}

The comparison run is against \textbf{HOV-SG~\cite{werby2024hovsg}}. We reimplement its published room-recovery construction (clearance thresholding + flooding) and run it as a full system, scored by the identical harness on the identical top-down maps (Sec.~\ref{sec:res-quant}). Our reimplementation departs from the published description in one respect: HOV-SG~\cite{werby2024hovsg} forms its wall and footprint rasters from two independent histograms, each taking its own data range, so the two cover slightly different physical extents before being combined; we rasterise both on a single grid. This removes a source of misregistration and so strengthens the baseline.

Within the room-detection pipeline itself, two internal design choices are ablated (Sec.~\ref{sec:res-ablation}): the seed-placement rule (Sec.~\ref{sec:seeds}) and the object-assignment rule (Sec.~\ref{sec:assign}). Each variant shares every other stage with the final method, so the difference measured is attributable to the ablated choice alone.


\section{Results}
\label{sec:results}



\begin{table}[t]
\caption{Room segmentation and object-to-room assignment over 10 floors of 6 HM3D-Semantics scenes, on the top-down grid (Sec.~\ref{sec:metrics}). P/R/$F_1$ micro-averaged; mIoU macro-averaged. Rooms: the 72 densely observed regions. Objects: all 2264 instances in 105 regions. 
}
\label{tab:rooms-main}
\centering
\vspace{2pt}
\small
\setlength{\tabcolsep}{4pt}
\setlength{\aboverulesep}{0pt}\setlength{\belowrulesep}{0pt}
\setlength{\extrarowheight}{0.7ex}
\begin{fittable}
\begin{tabular}{lccccccccccc}
\toprule
& & & \multicolumn{6}{c}{Rooms (72 observed regions)}
  & \multicolumn{3}{c}{Objects (2264 instances)} \\
\cmidrule(lr){4-9}\cmidrule(l){10-12}
Method & $n_{\mathrm{pred}}$ & mIoU$\uparrow$
& P@.25$\uparrow$ & R@.25$\uparrow$ & F1@.25$\uparrow$ & P@.5$\uparrow$ & R@.5$\uparrow$ & F1@.5$\uparrow$
& Acc.$\uparrow$ & ARI$\uparrow$ & NMI$\uparrow$ \\
\midrule
HOV-SG~\cite{werby2024hovsg} & 44 & 0.734
& \textbf{0.977} & 0.597 & 0.741 & \textbf{0.909} & 0.556 & 0.690
& 0.684 & 0.488 & 0.593 \\
\ourrow \textbf{Ours} & 74 & \textbf{0.802}
& 0.878 & \textbf{0.903} & \textbf{0.890} & 0.811 & \textbf{0.833} & \textbf{0.822}
& \textbf{0.801} & \textbf{0.696} & \textbf{0.744} \\
\bottomrule
\end{tabular}
\end{fittable}
\vspace{2pt}
{\small $|\Delta m| = |n_{\mathrm{pred}} - 72|$: HOV-SG~\cite{werby2024hovsg}: 28; \textbf{Ours}: \textbf{2}.}
\end{table}

\subsection{Quantitative Comparison}
\label{sec:res-quant}
Table~\ref{tab:rooms-main} compares our final configuration against our reimplementation of HOV-SG~\cite{werby2024hovsg}on identical top-down maps. Room partitioning is scored over the 72 annotated regions that the walk observes densely enough to admit a geometric comparison; object assignment is scored over all 2264 instances and the 105 regions they inhabit, since an object annotated inside a sparsely observed region is still a real object with a correct room. Per-floor figures are given in Table~\ref{tab:scenewise}. Published numbers for the closest prior systems, which are context rather than a controlled comparison, are placed beside ours in Appendix~\ref{app:published} and Table~\ref{tab:published}.


\paragraph{What is and is not established.}
A paired Wilcoxon signed-rank test over the ten floors of Table~\ref{tab:scenewise} separates the two methods on object assignment, with accuracy, ARI and NMI all at $p = 0.002$ and ours ahead on every floor. It does not separate them on room mIoU, where the $+0.068$ margin holds on 7 of 10 floors and gives $p = 0.065$. We therefore treat the room geometry as a consistent trend at this sample size and rest the comparison on object assignment, which is both the better-powered measurement and the quantity a scene graph is queried for. Ten floors is a small paired sample and we claim no more from it than the tests support.

Because both methods use the identical assignment rule for the object columns of Table~\ref{tab:rooms-main}, the accuracy gap ($0.801$ against $0.684$) and the ARI gap ($0.696$ against $0.488$) are attributable to room segmentation rather than to how objects are placed once rooms exist.\footnote{The rule is identical including its fallback: an object whose entire extent misses every labelled cell takes the geodesically nearest room. This resolves $25.1\%$ of objects under our segmentation and $21.0\%$ under HOV-SG's, so the comparison is between two segmentations under one shared procedure.}

\begin{table}[!t]
\caption{Overlap precision and recall under the Bormann/Hydra criterion
(Eq.~\eqref{eq:pr}) on our six HM3D-Semantics scenes, over grid cells (primary,
Sec.~\ref{sec:metrics}) and surface points. The published block is context
rather than a controlled comparison, since those systems consume posed RGB-D or
GT poses with simulator depth and average over 3D voxels per scene
(Sec.~\ref{sec:baselines}); LEXI-SG is nearest to our input setting. Our
reimplementation brackets HOV-SG's published precision of $0.87$, giving $0.84$
over cells and $0.77$ over points, which is our evidence that the port is
faithful. Ranking is \emph{within} each block, since voxel, surface-point and
grid-cell scores are not commensurable: \textbf{bold} = best, \snd{underline} =
second-best; the two-method blocks carry no second-best mark. $\ddag$: HOV-SG's
authors' evaluation of Hydra. \xmark: no valid reconstruction.}
\label{tab:published}
\centering
\vspace{1em}
\small
\setlength{\tabcolsep}{4pt}
\setlength{\aboverulesep}{0pt}\setlength{\belowrulesep}{0pt}
\setlength{\extrarowheight}{0.5ex}
\begin{fittable}
    \begin{tabular}{l|lccccccc}
    \toprule
    \multicolumn{2}{c}{} & \multicolumn{6}{c}{HM3D-Semantics scene} & \\
    \cmidrule(lr){3-8} & Method & 824 & 829 & 843 & 873 & 877 & 890 & Avg. \\
    \midrule
    \multirow{11}{*}{\rotatebox{90}{Precision}} & \multicolumn{7}{l}{\emph{published, 3D free-space voxels}} \\
     & \quad Hydra~\cite{hughes2022hydra}$^{\ddag}$ & \snd{0.80} & \snd{0.86} & \snd{0.87} & \textbf{0.97} & \textbf{0.82} & \textbf{0.95} & \textbf{0.88} \\
     & \quad HOV-SG~\cite{werby2024hovsg} & \textbf{0.81} & \textbf{0.89} & \textbf{0.89} & \snd{0.96} & \snd{0.75} & \textbf{0.95} & \snd{0.87} \\
     & \quad LEXI-SG (GT poses) & 0.56 & 0.75 & 0.77 & 0.78 & 0.59 & \snd{0.81} & 0.71 \\
     & \quad LEXI-SG & 0.52 & 0.58 & 0.77 & \xmark & 0.54 & 0.62 & --- \\
    \cmidrule(l){2-9} & \multicolumn{7}{l}{\emph{surface points}} \\
     & \quad HOV-SG~\cite{werby2024hovsg}& 0.79 & 0.76 & 0.67 & 0.79 & 0.70 & 0.86 & 0.77 \\
     & \oc \quad \textbf{Ours} & \oc \textbf{0.88} & \oc \textbf{0.81} & \oc \textbf{0.80} & \oc \textbf{0.84} & \oc \textbf{0.83} & \oc \textbf{0.88} & \oc \textbf{0.84} \\
    \cmidrule(l){2-9} & \multicolumn{7}{l}{\emph{grid cells (primary)}} \\
     & \quad HOV-SG~\cite{werby2024hovsg}& 0.89 & 0.88 & 0.71 & 0.86 & 0.75 & 0.93 & 0.84 \\
     & \oc \quad \textbf{Ours} & \oc \textbf{0.97} & \oc \textbf{0.93} & \oc \textbf{0.87} & \oc \textbf{0.94} & \oc \textbf{0.93} & \oc \textbf{0.95} & \oc \textbf{0.93} \\
    \midrule
    \multirow{11}{*}{\rotatebox{90}{Recall}} & \multicolumn{7}{l}{\emph{published, 3D free-space voxels}} \\
     & \quad Hydra~\cite{hughes2022hydra}$^{\ddag}$ & 0.78 & 0.85 & \snd{0.79} & \snd{0.80} & \snd{0.89} & 0.63 & 0.79 \\
     & \quad HOV-SG~\cite{werby2024hovsg} & 0.80 & 0.88 & \textbf{0.87} & 0.68 & \textbf{0.92} & \snd{0.88} & \snd{0.84} \\
     & \quad LEXI-SG (GT poses) & \textbf{0.96} & \snd{0.91} & 0.74 & \textbf{0.85} & 0.78 & \textbf{0.93} & \textbf{0.86} \\
     & \quad LEXI-SG & \snd{0.85} & \textbf{0.98} & 0.69 & \xmark & 0.86 & 0.74 & --- \\
    \cmidrule(l){2-9} & \multicolumn{7}{l}{\emph{surface points}} \\
     & \quad HOV-SG~\cite{werby2024hovsg}(reimpl.) & 0.74 & \textbf{0.89} & \textbf{0.88} & \textbf{0.88} & \textbf{0.90} & \textbf{0.93} & \textbf{0.87} \\
     & \oc \quad \textbf{Ours} & \oc \textbf{0.84} & \oc 0.85 & \oc 0.85 & \oc 0.86 & \oc 0.81 & \oc 0.87 & \oc 0.84 \\
    \cmidrule(l){2-9} & \multicolumn{7}{l}{\emph{grid cells (primary)}} \\
     & \quad HOV-SG~\cite{werby2024hovsg}(reimpl.) & 0.89 & \textbf{0.97} & \textbf{0.94} & \textbf{0.93} & \textbf{0.94} & \textbf{0.97} & \textbf{0.94} \\
     & \oc \quad \textbf{Ours} & \oc \textbf{0.93} & \oc 0.92 & \oc 0.90 & \oc \textbf{0.93} & \oc 0.88 & \oc 0.92 & \oc 0.91 \\
    \bottomrule
    \end{tabular}
\end{fittable}
\end{table}

\subsection{Qualitative Comparison}
\label{sec:res-qual}

\begin{figure}[!t]
\centering
\resizebox{\columnwidth}{!}{
    \newcommand{\qualcell}[1]{\raisebox{-0.5\height}{\includegraphics[width=0.30\columnwidth]{#1}}}
    \begin{tabular}{cccc}
    & {HOV-SG} & {Ours} & {Ground Truth} \\[1pt]
    \rotatebox[origin=c]{90}{\texttt{00829} f0} & 
    \qualcell{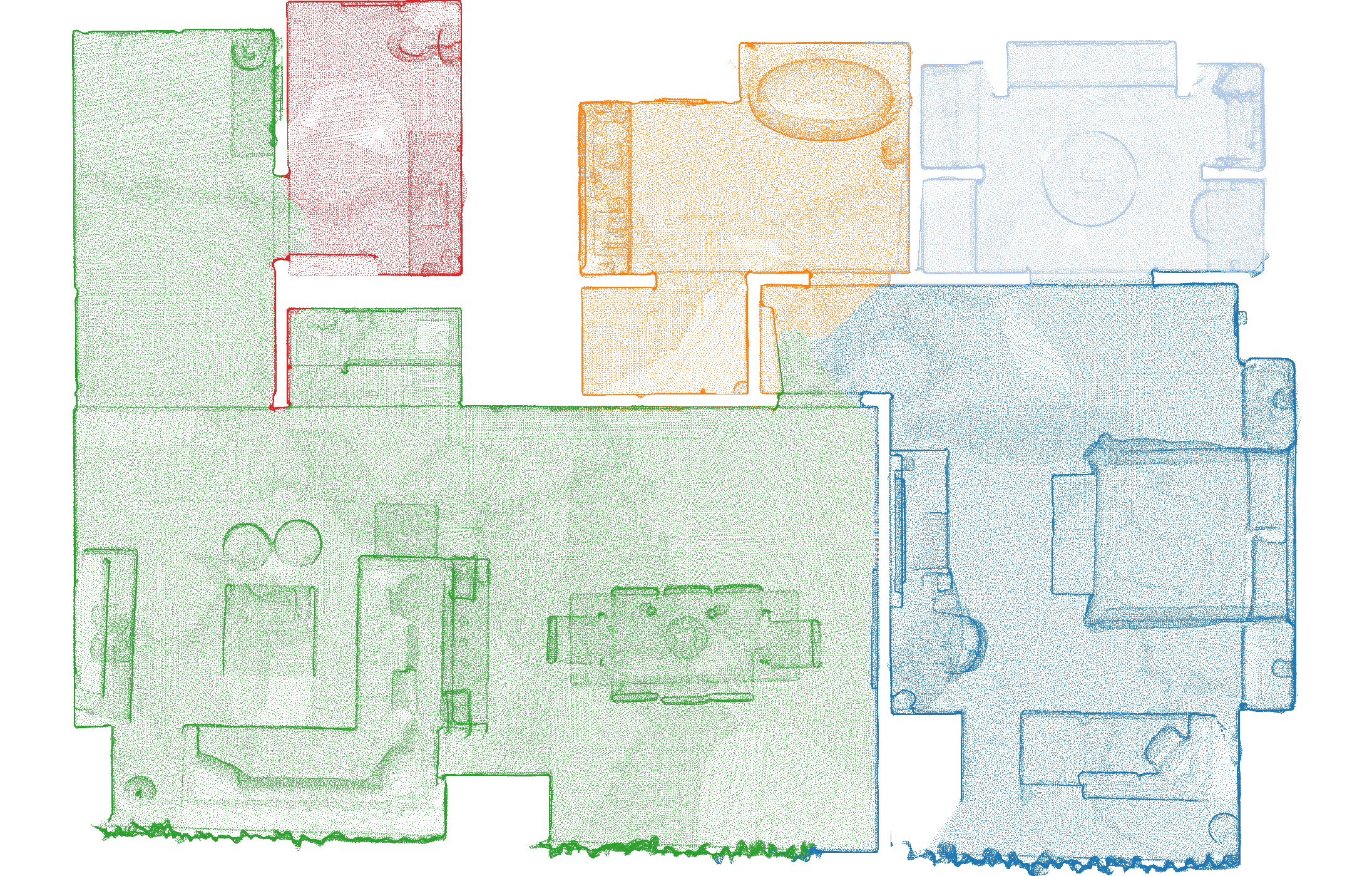} & \qualcell{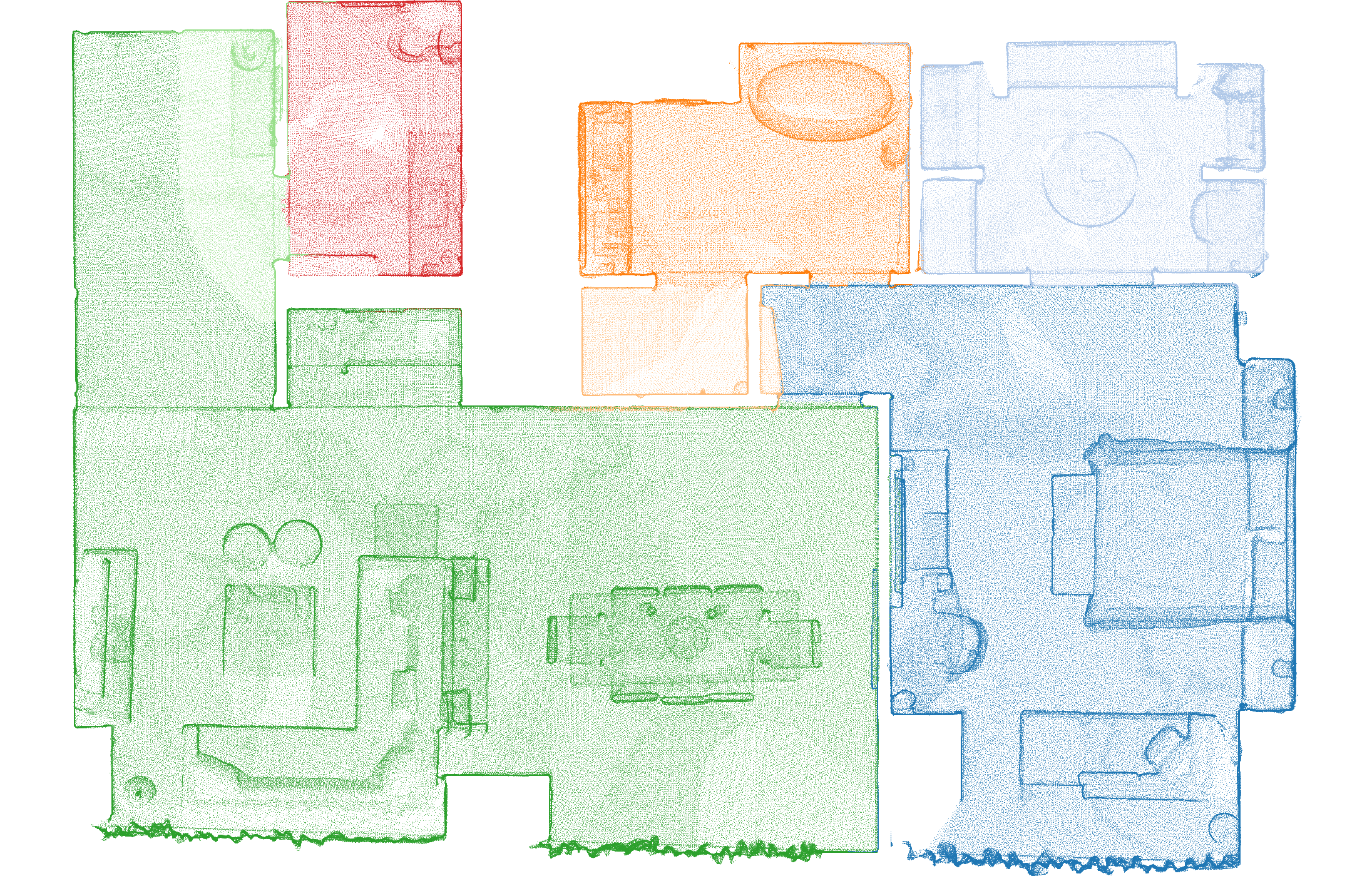} &
    \qualcell{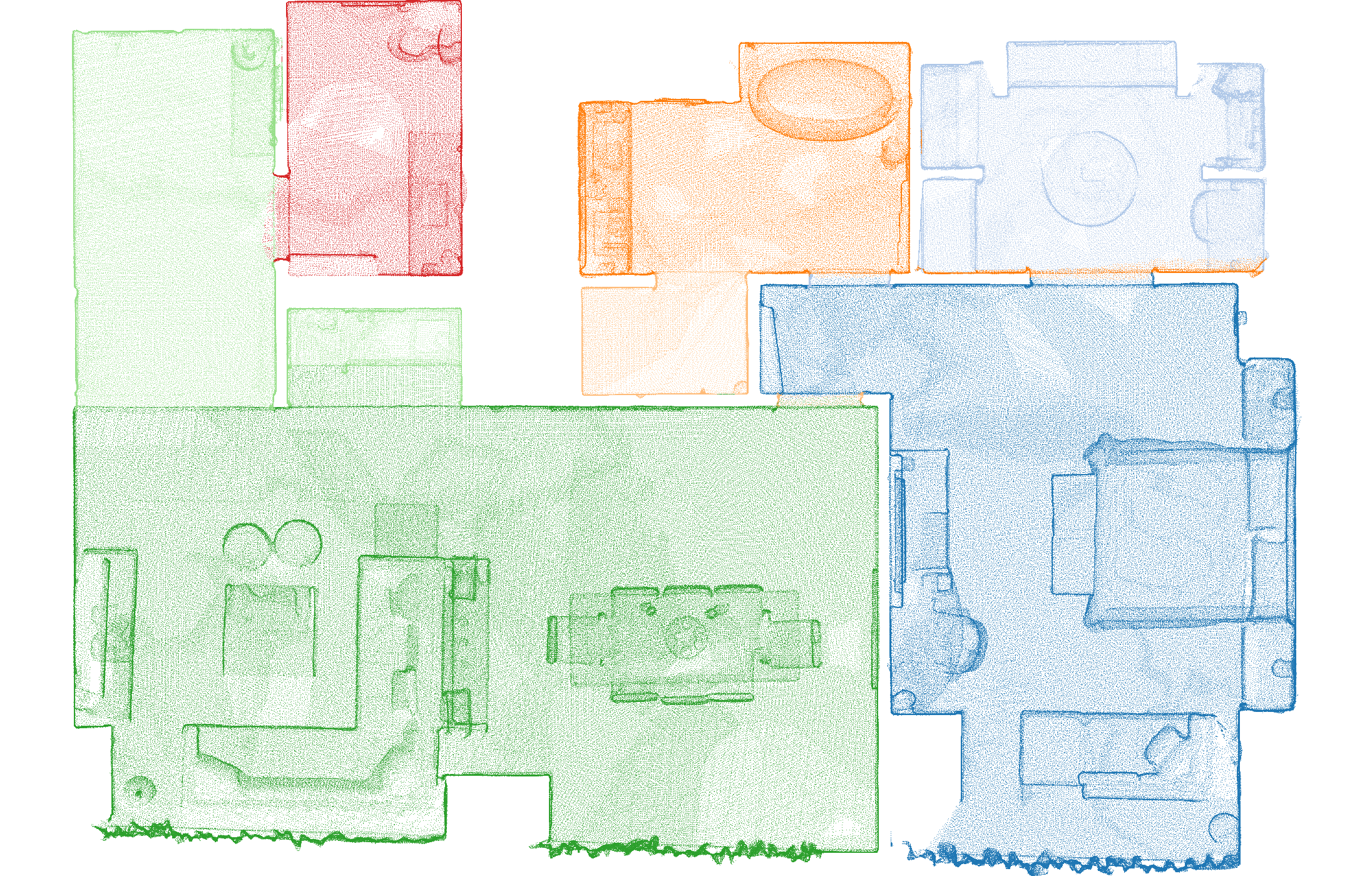} \\
    & 0.842 & 0.795 \\\hline
    \rotatebox[origin=c]{90}{\texttt{00843} f0} & \qualcell{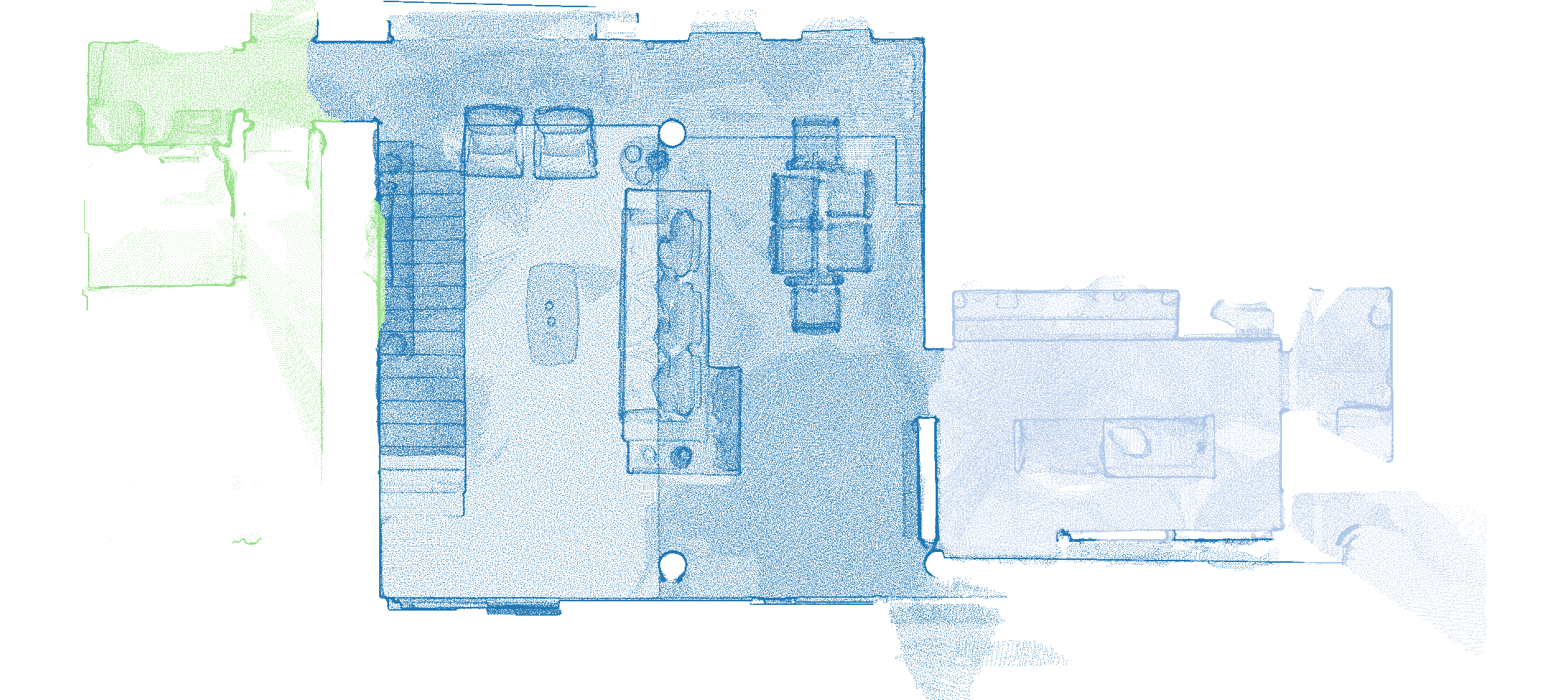} & \qualcell{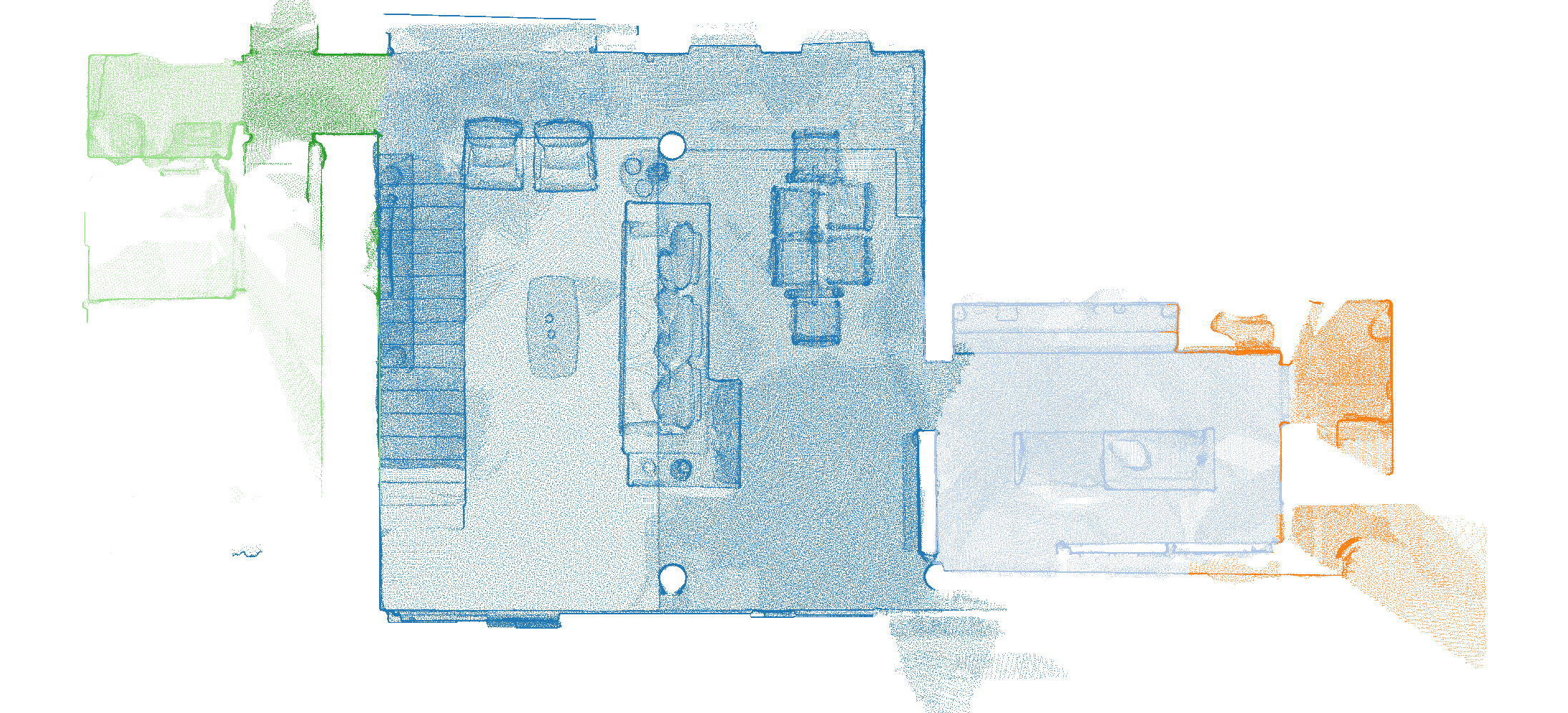} & \qualcell{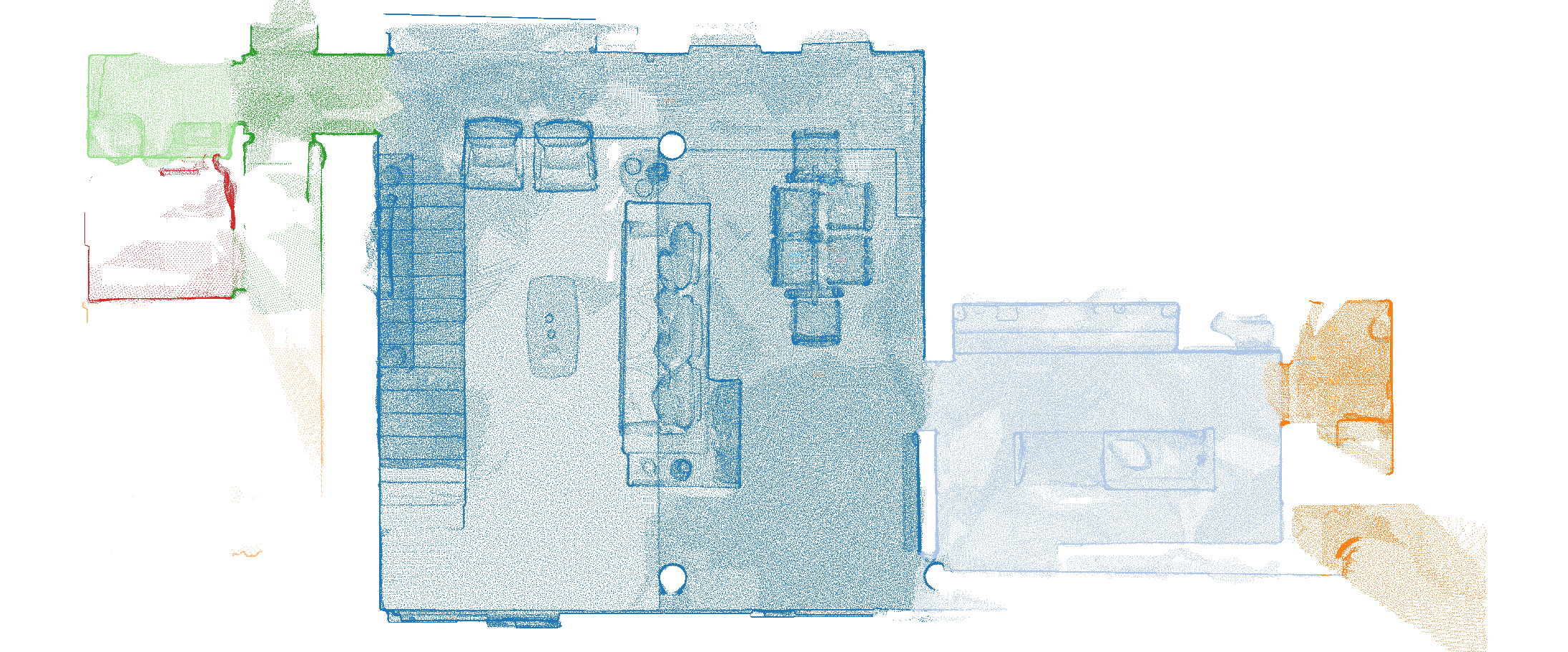} \\
    & 0.611 & 0.734 \\\hline
    \rotatebox[origin=c]{90}{\texttt{00877} f0} & \qualcell{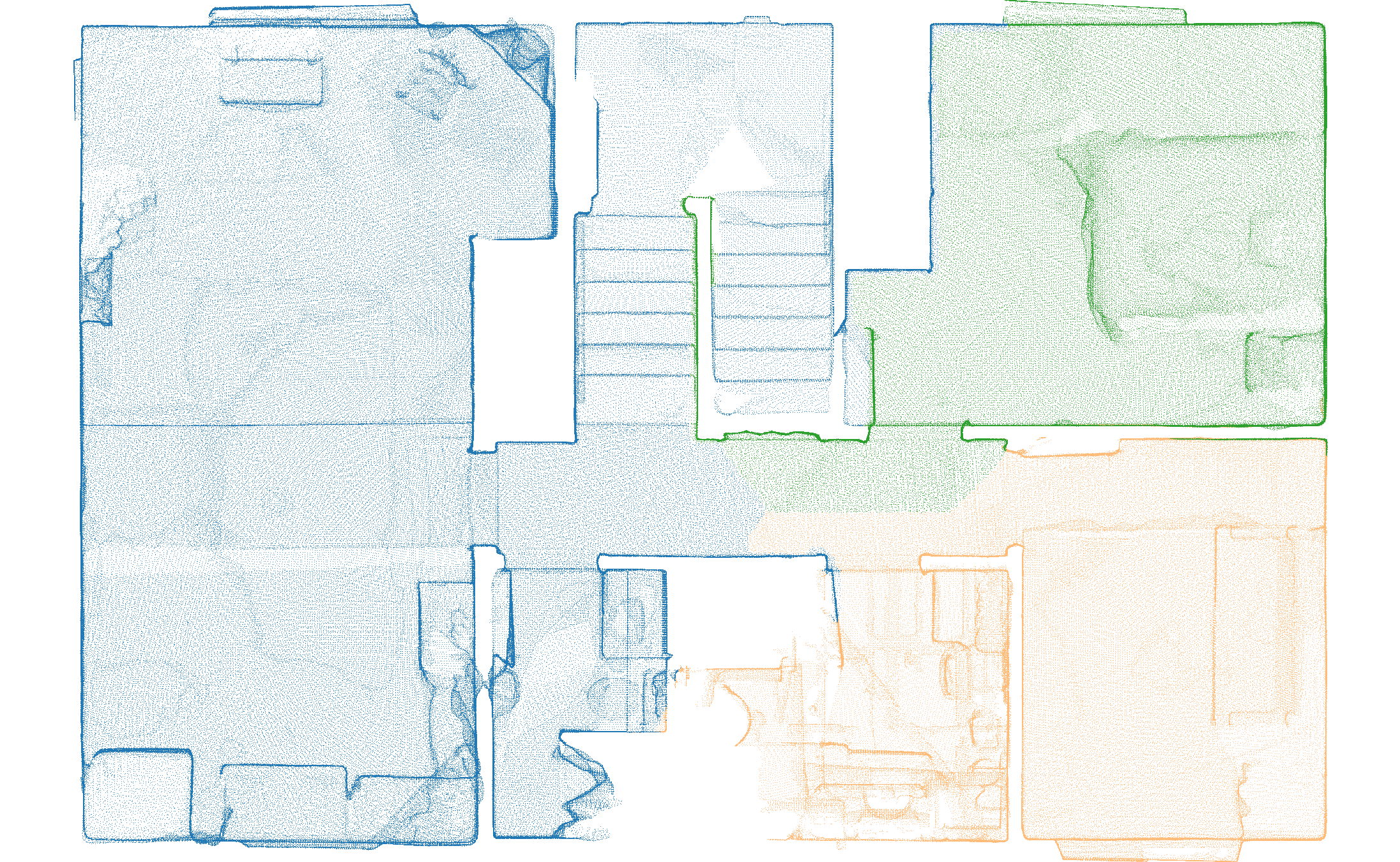} & \qualcell{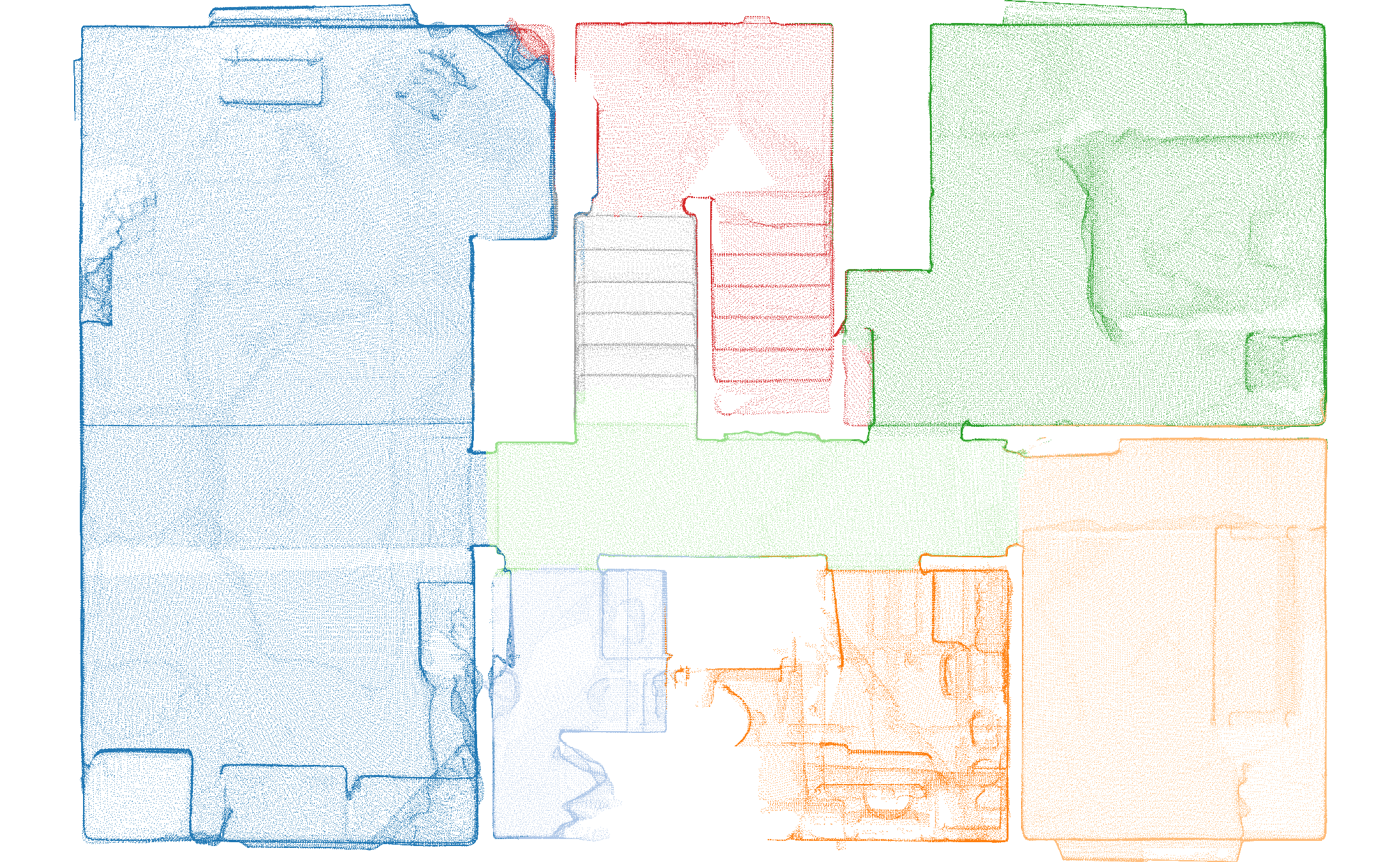} & \qualcell{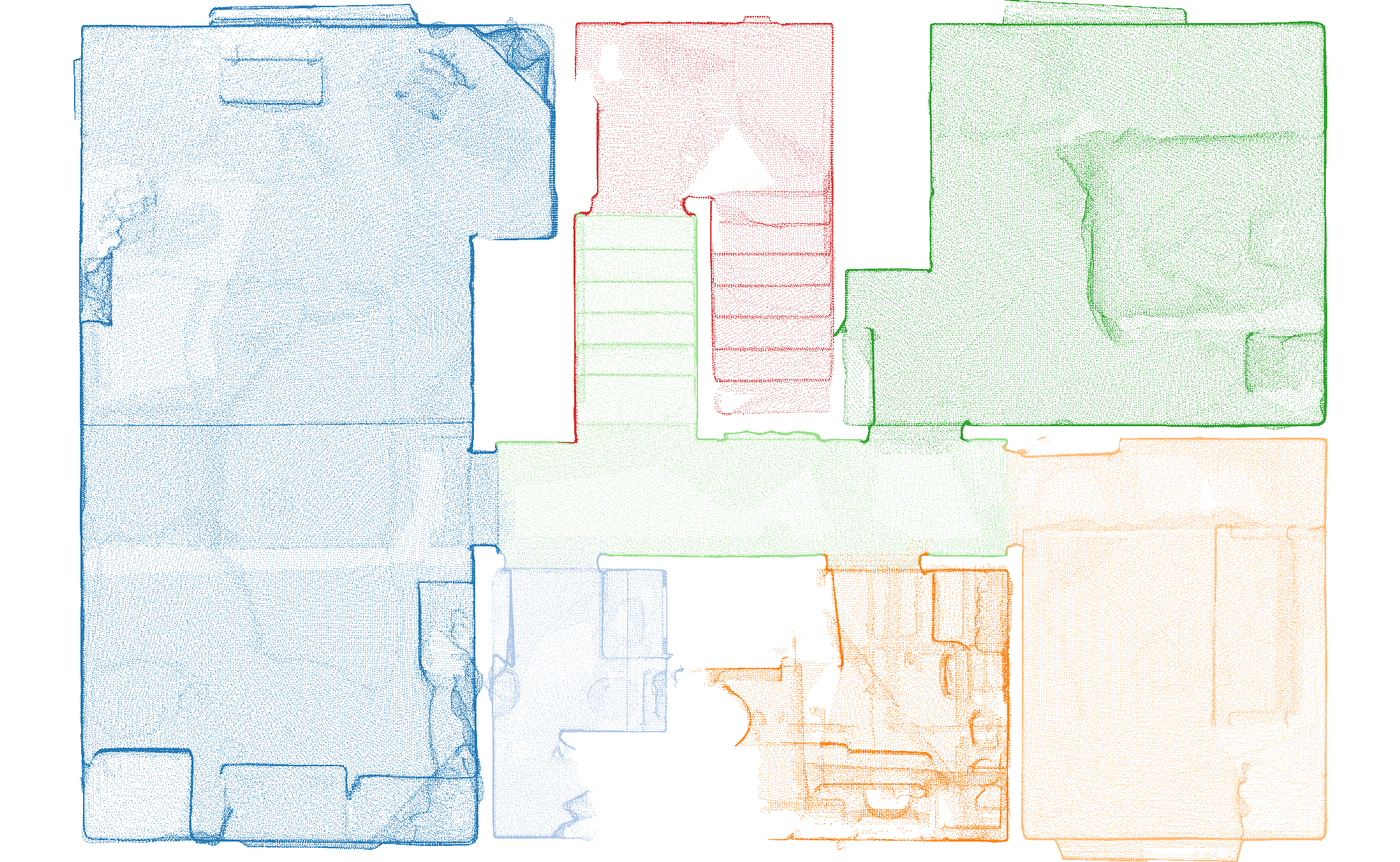} \\
    & 0.698 & 0.859 \\\hline
    \rotatebox[origin=c]{90}{\texttt{00873} f0} & \qualcell{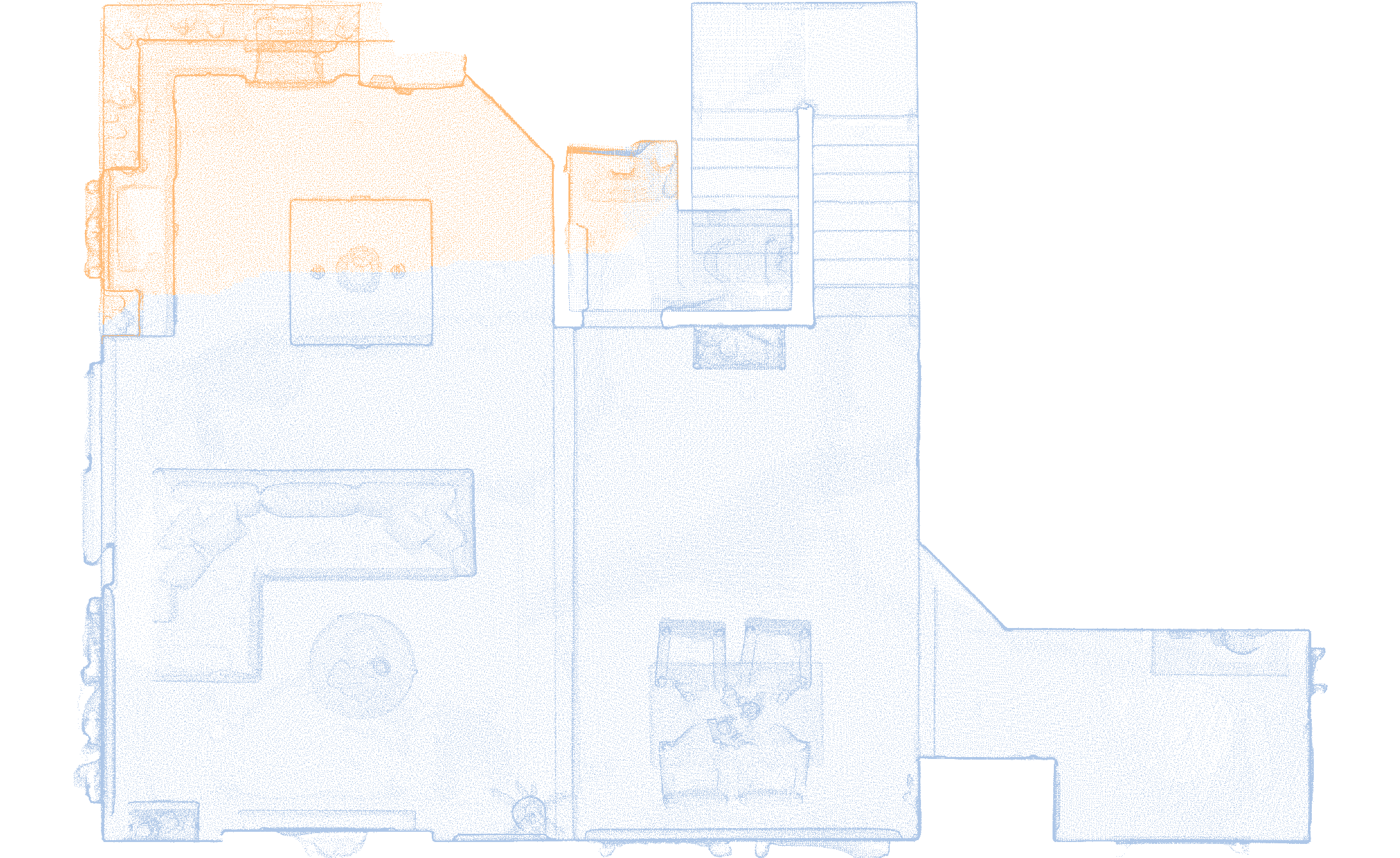} & \qualcell{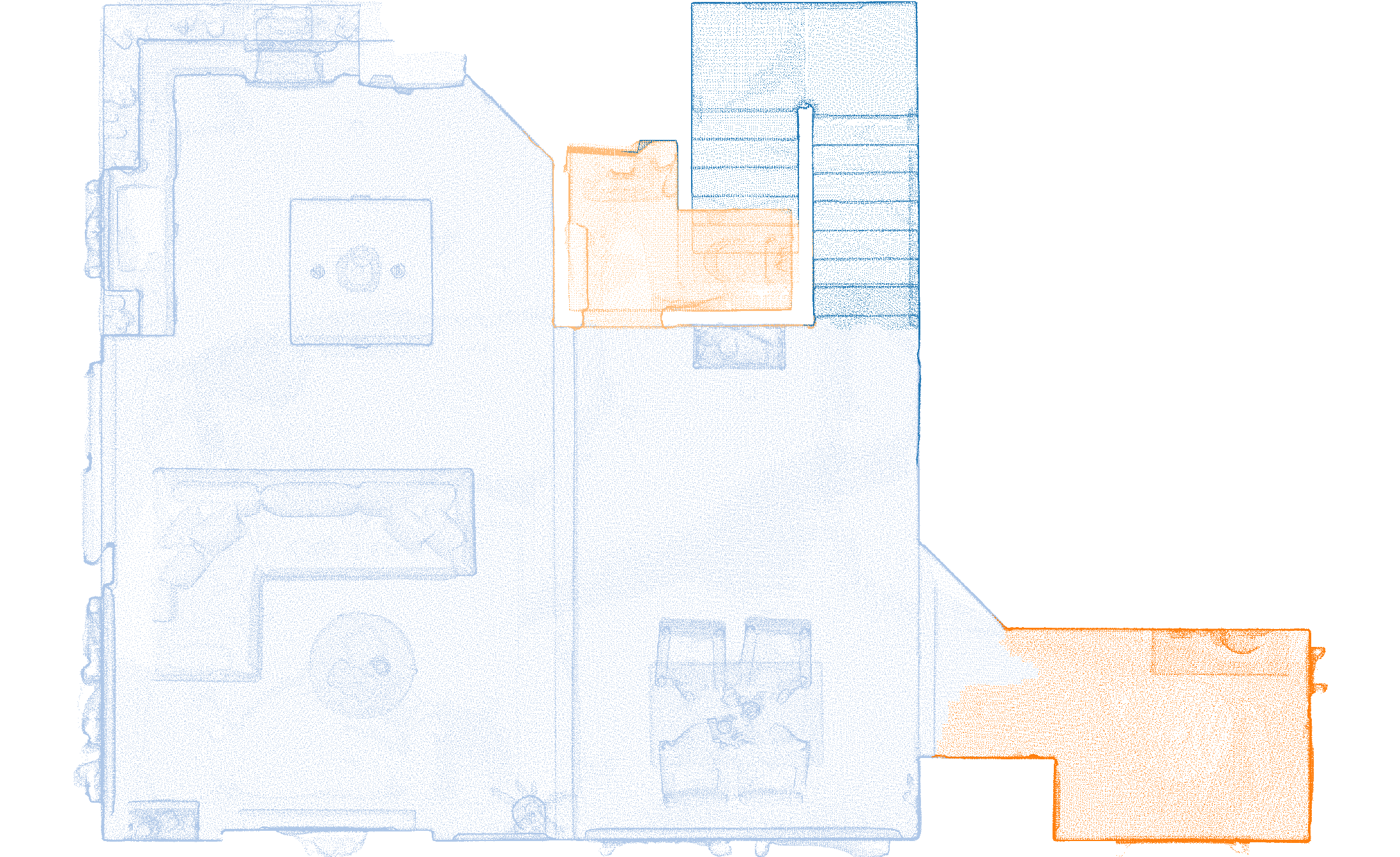} & \qualcell{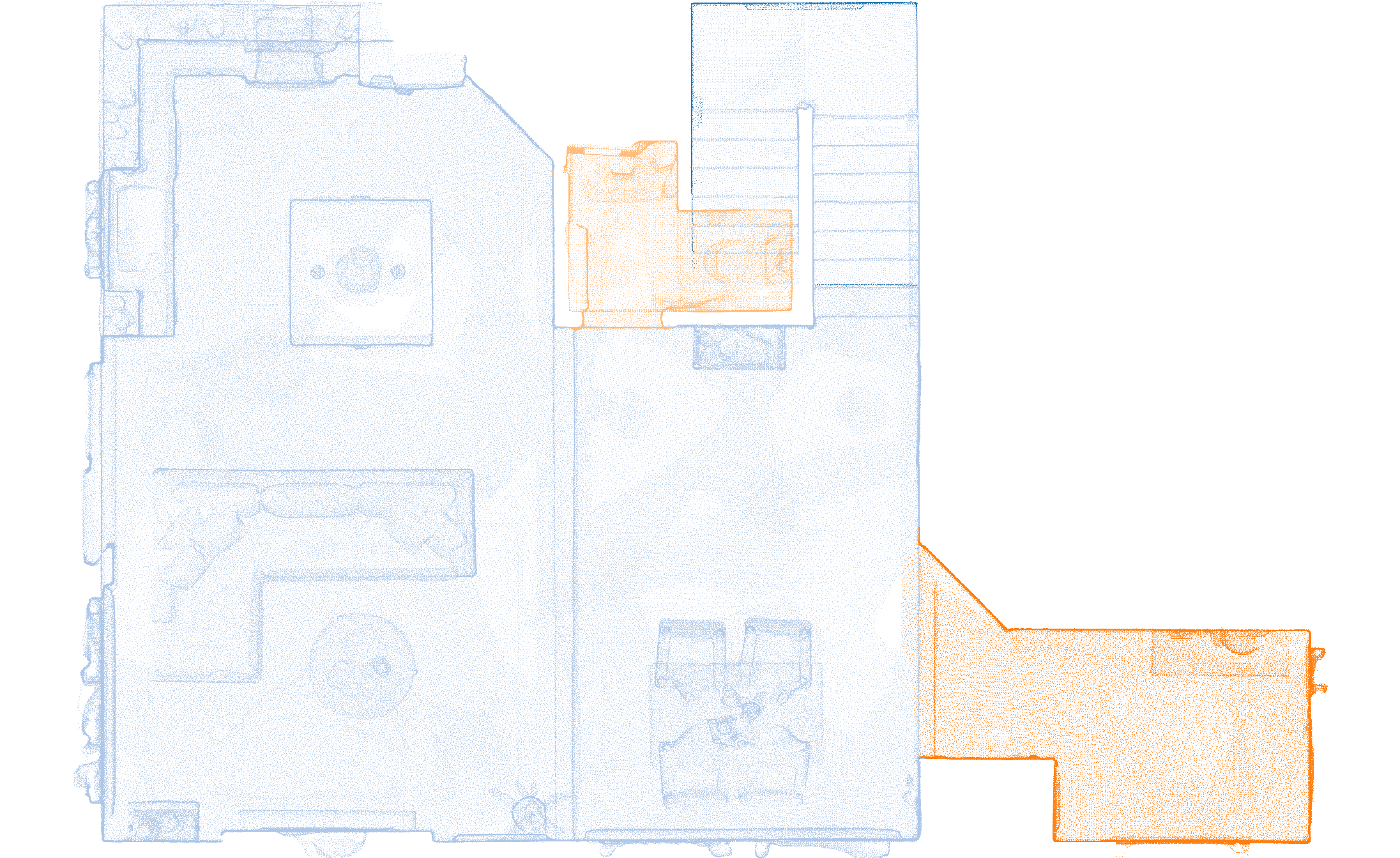} \\
    & 0.350 & 0.647 \\\hline
    \rotatebox[origin=c]{90}{\texttt{00843} f1} & \qualcell{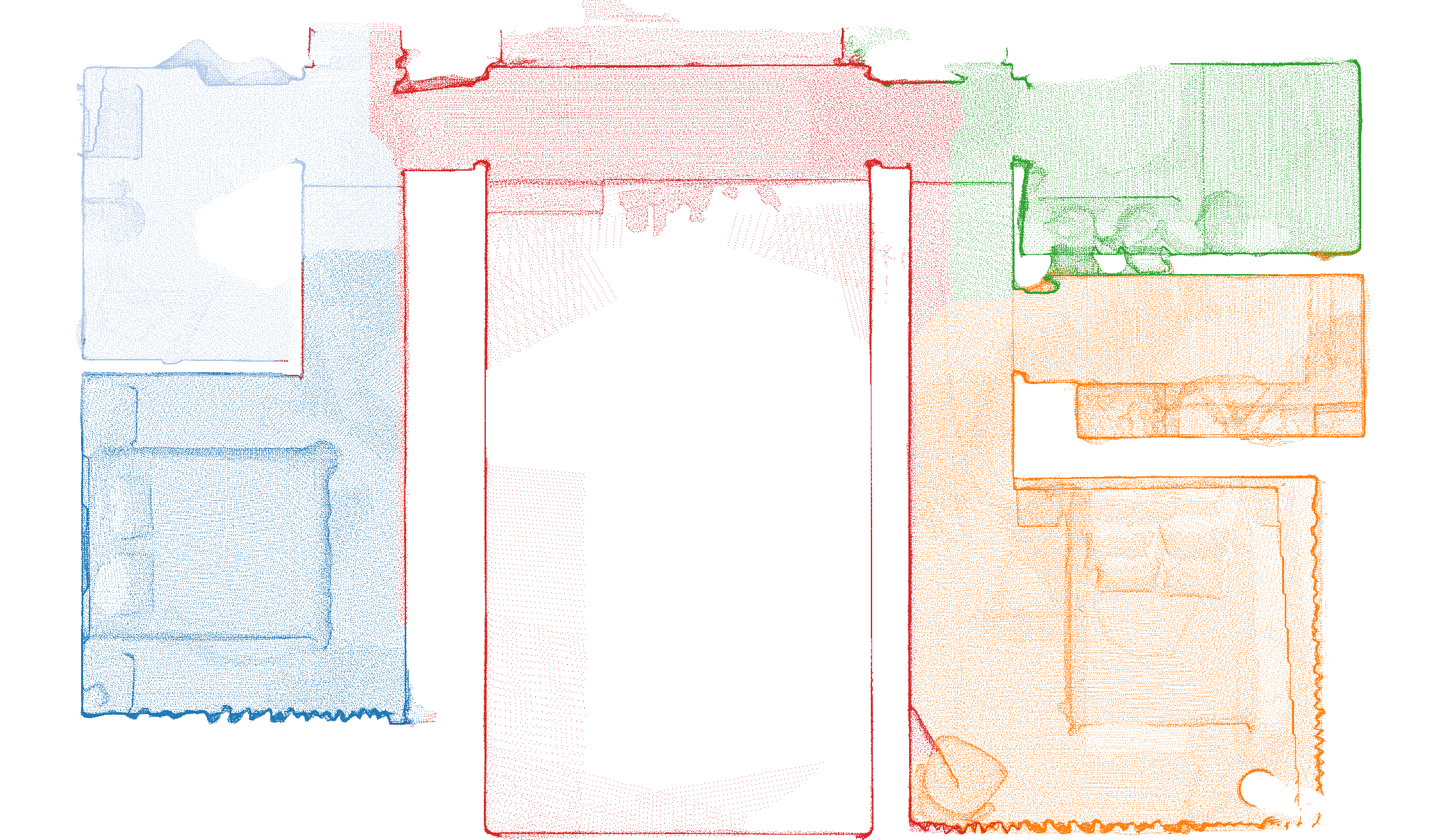} & \qualcell{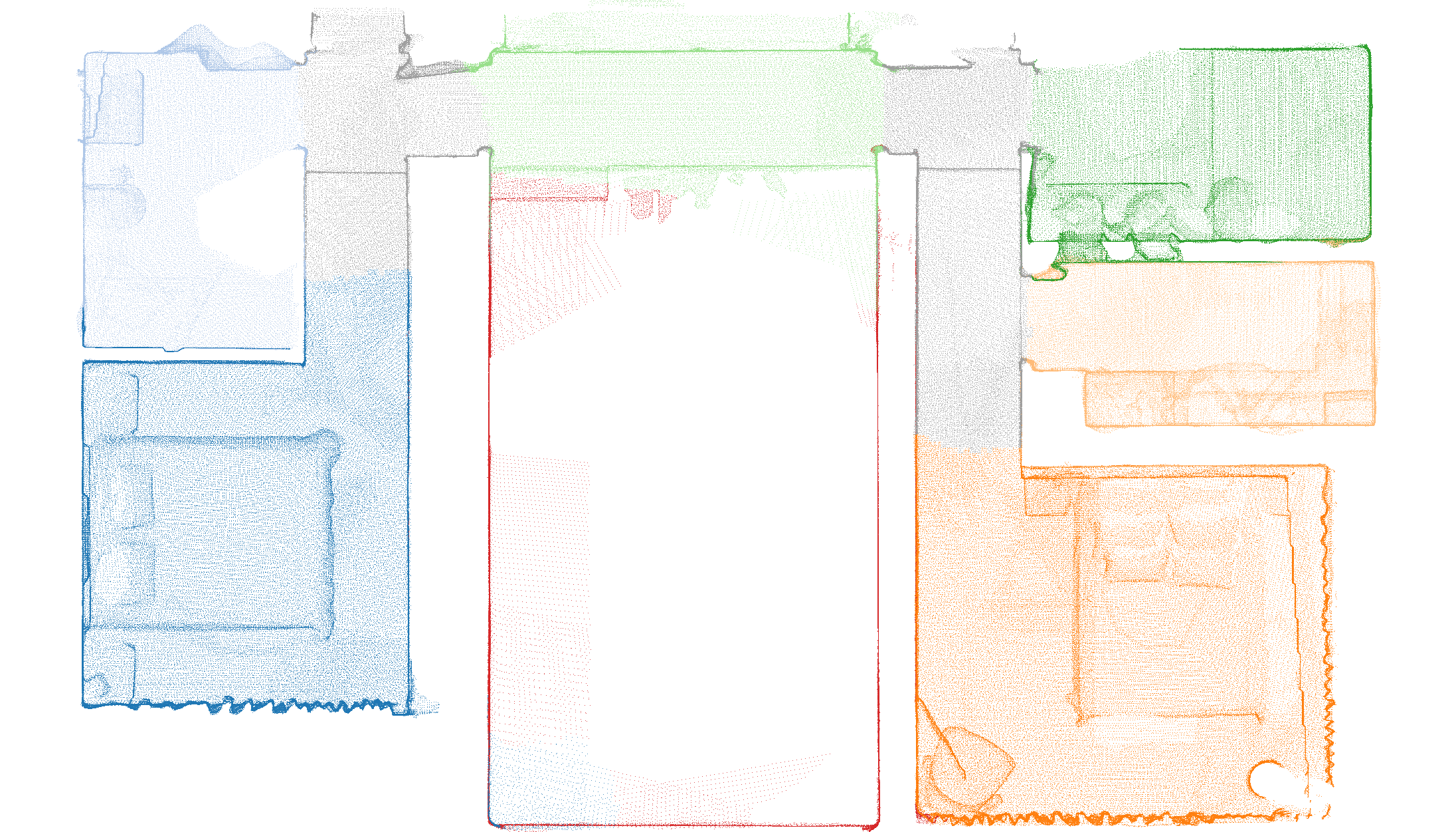} & \qualcell{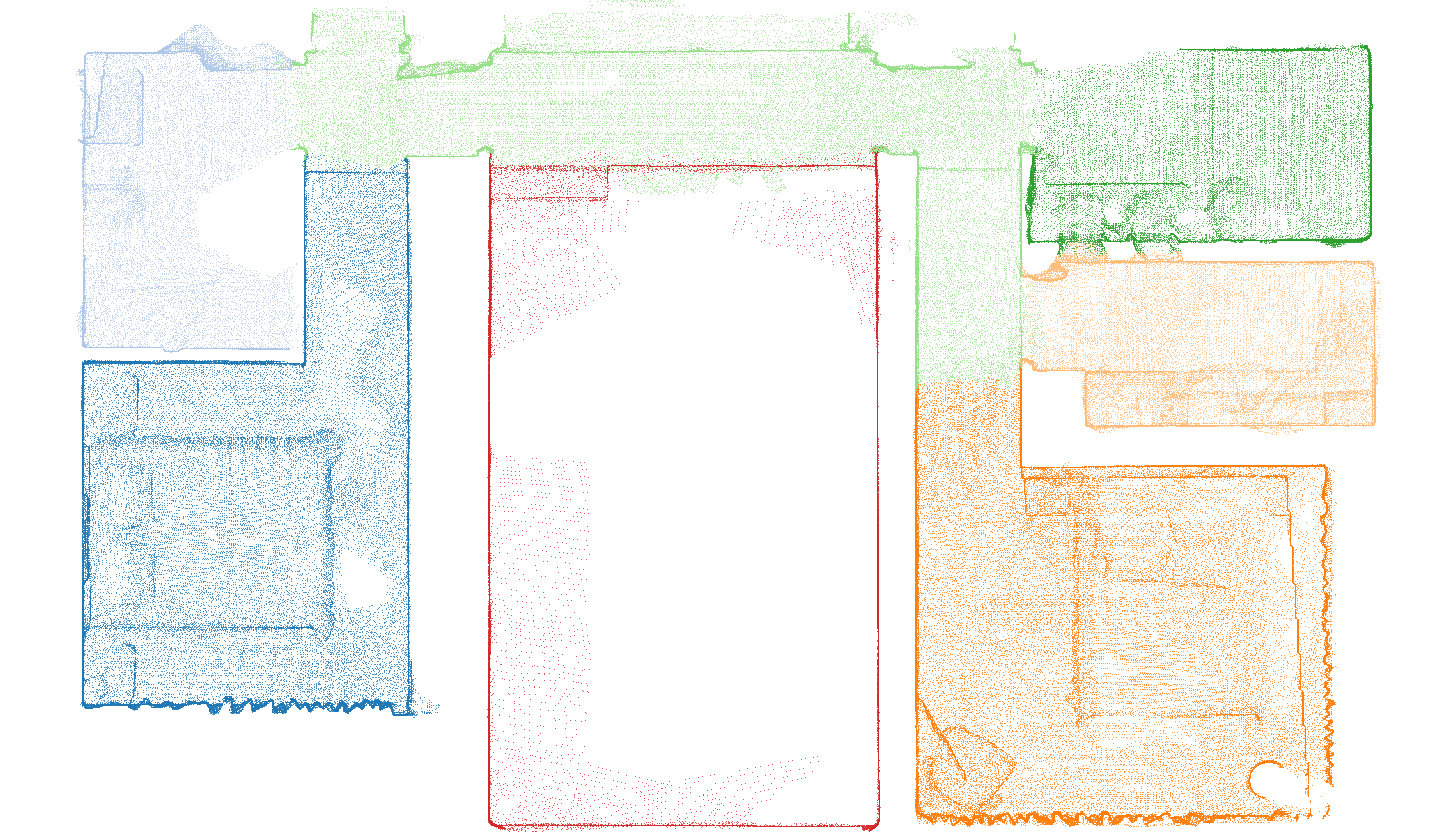} \\
    & 0.692 & 0.804 \\
    \end{tabular}
}


\caption{Qualitative room segmentation across three scenes, one sequence per row, left to right: HOV-SG, ours, ground truth. Mean IoU for HOV-SG~\cite{werby2024hovsg}and our method is reported below each prediction
}
\label{fig:qualitative}
\end{figure}

Figure~\ref{fig:qualitative} places our reimplementation of HOV-SG~\cite{werby2024hovsg}, our prediction and the ground truth on the same grid for five floors spanning the range of outcomes, from \texttt{00873} floor~0, where the margin is largest (mIoU $0.350$ against $0.647$), to \texttt{00829} floor~0, one of three floors where HOV-SG~\cite{werby2024hovsg} scores higher ($0.842$ against $0.795$, Table~\ref{tab:scenewise}).

One pattern holds across all five rows. HOV-SG~\cite{werby2024hovsg} under-segments, recovering 5, 3, 3, 2 and 5 rooms against 7, 7, 7, 4 and 7 observed regions, so spaces the annotation separates are merged into one. We recover 7, 5, 8, 4 and 9, matching the region count exactly on \texttt{00829} floor~0 and \texttt{00873} floor~0.

Where we score lower the cause is not the room count but the boundary. On \texttt{00829} floor~0 we recover 7 rooms for 7 regions and still fall below HOV-SG~\cite{werby2024hovsg}on mIoU, because several of our boundaries stop short of the annotated ones and the rooms overlap their regions loosely. Mean IoU rewards a smaller number of well-fitted rooms over a larger number of approximate ones, and that is the trade this figure makes visible.

The objects on that floor behave differently. \texttt{00829} floor~0 gives object accuracy $0.847$ and ARI $0.752$, among the better floors of the ten, so a boundary that fits its region loosely still contains the objects inside it. This is the room-object disagreement Sec.~\ref{sec:metrics} anticipates. The residual errors fall where Sec.~\ref{sec:assign} predicts: an object straddling a boundary has its extent split between two rooms, so the majority is decided by a thin margin, and the nearest-room fallback resolves 53 of the floor's 196 instances.

\subsection{Ablation Study}
\label{sec:res-ablation}

We ablate the three stages for which a rejected alternative was run end to end: seed placement (Sec.~\ref{sec:seeds}), the doorway seal (Sec.~\ref{sec:cleanup}) and the object-to-room assignment rule (Sec.~\ref{sec:assign}). In each case the two variants share every other stage of the pipeline, so the measured difference is attributable to the ablated choice alone. Table~\ref{tab:ablation} reports the results.

\begin{table}[!t]
\caption{Ablations. Each stage is scored on the metric it affects: seeding and
the doorway seal by room mIoU over the 72 regions of
Table~\ref{tab:rooms-main}, the assignment rule by object ARI over the 2264
instances. The variants within an ablation share every other stage, so the
difference is attributable to the ablated choice. $p$: paired two-sided
Wilcoxon over the 10 floors, with the win/loss record beside it; $\dag$: does
not reach $p<0.05$, reported as a trend. Bold: the variant used.}
\label{tab:ablation}
\centering
\vspace{1em}
\small
\setlength{\tabcolsep}{4pt}
\begin{fittable}
    \begin{tabular}{lccc}
    \toprule
    Variant & $n_{\mathrm{pred}}$ & Value & $p$ (W/L) \\
    \midrule
    \multicolumn{4}{l}{\emph{Seed placement (room mIoU)}} \\
    \quad Farthest-point sampling & 75 & 0.781 & \multirow{2}{*}{0.064$^{\dag}$ (9/1)} \\
    \quad Camera trajectory & 74 & \textbf{0.802} & \\
    \midrule
    \multicolumn{4}{l}{\emph{Doorway seal (room mIoU)}} \\
    \quad Off & 71 & 0.768 & \multirow{2}{*}{\textbf{0.002} (10/0)} \\
    \quad On & 74 & \textbf{0.802} & \\
    \midrule
    \multicolumn{4}{l}{\emph{Assignment rule (object ARI)}} \\
    \quad Visibility agreement & 74 & 0.689 & \multirow{2}{*}{0.193$^{\dag}$ (7/3)} \\
    \quad Surface majority & 74 & \textbf{0.696} & \\
    \bottomrule
    \end{tabular}
\end{fittable}
\end{table}



\subsection{Limitations}
\label{sec:res-fail}

\paragraph{Wide openings.}
\texttt{00873-bxsVRursffK} floor~0 is the lowest-scoring floor of the ten (mIoU $0.647$, Table~\ref{tab:scenewise}). It is substantially open-plan, and the failure is the one Sec.~\ref{sec:closure} anticipates: a space joined to its neighbour by an opening comparable to its own extent does not saturate before that neighbour, so the two seal together. Of the eleven annotated regions on this floor only four are observed densely enough to enter the room evaluation, and we recover four rooms against them; the remaining seven are absorbed. Object accuracy is nonetheless $0.846$, because one region holds a large share of the floor's 175 instances and maps cleanly onto the merged room, while the absorbed regions hold one or two objects each. ARI, which is not forgiving of that imbalance, falls to $0.590$. This is the room-object disagreement Sec.~\ref{sec:metrics} anticipates: the boundary is wrong, yet conditioned on the wrong boundary the objects are placed consistently with it.


\subsection{Discussion}
\label{sec:res-discussion}

Three limits on what these results establish should be stated plainly. The object-assignment comparison separates the two methods on every floor, but the room comparison does not ($p = 0.065$), so we describe the room figures as a consistent trend and rest the claim on object assignment. Trajectory seeding is likewise a trend rather than a result: it leads on nine of ten floors but $p = 0.064$, and its firmer advantages are procedural, in requiring no eligible pool and no random choice. And while the method is motivated by the argument that a per-region closure scale is preferable to one global radius, the experiments here compare against a published room-recovery construction rather than against a single global closing operator, so that motivation rests on the mechanism and on the comparison to HOV-SG, not on a controlled sweep over closure radii. Such a sweep on the same rasterised maps would test it directly and is the most valuable single addition to this evaluation.

\paragraph{Recurring failure modes.}
\label{sec:failuretax}
Items~1 to~3 affect every result reported here. Items~4 and~5 arise only in the monocular pipeline, where the vertical is estimated and objects are tracked, and cannot occur in the quantitative evaluation, whose vertical is given by the dataset and whose objects are annotated instances.

\section{Conclusion}
\label{sec:conclusion}
We recover the room layer of an indoor scene graph from pose-free monocular video, where the top-down boundary is broken by the trajectory as much as by the architecture. The method treats every break by its width rather than its cause: rooms are declared as free-space regions saturate under progressive boundary thickening, each at its own closure scale, and objects are placed by the room holding most of their reconstructed extent rather than by containment of a single point, so an object near a boundary is decided by a margin that narrows rather than by a test that flips. Neither stage classifies an opening as a door.

On ten floors of six HM3D-Semantics scenes, objects are assigned to the correct room for 0.801 of 2264 annotated instances against 0.684 for our reimplementation of the clearance-and-flooding construction (ARI 0.696 against 0.488), a difference significant over the ten floors under a paired test. The recovered partition is also closer to the annotated room count ($|\Delta m|{=}2$ against $28$) and scores higher $F_1$ at both overlap thresholds, though the mean-IoU margin alone does not reach significance at this sample size.

The method does not separate spaces joined by an opening comparable to their own extent, which is the open-plan case and admits no purely geometric criterion. Recovering metric scale, storey separation, and room adjacency from the same input remain open.

\bibliographystyle{unsrt}
\bibliography{references}

\clearpage

\begin{center}
{\Large\bfseries ProClosure: Hierarchical Room-Object Assignment using Progressive Boundary Closure from Monocular Video\\Supplementary Material - Appendix}
\end{center}
\vspace{1em}
\appendix
  \emergencystretch=1.2em


\section{Reconstruction and Structural Segmentation}
\label{sec:recon}
\label{app:recon}

\paragraph{Reconstruction.}
Geometry comes from a feed-forward pointmap SLAM front end \cite{mast3rslam}, which for each keyframe $I_t \in \mathcal{K}$ returns a dense per-pixel point map in a shared world frame, a per-pixel confidence, and the pose $T_t \in SE(3)$. Low-confidence pixels are discarded; the confident points of all keyframes are pooled, voxel-downsampled, and outlier-cleaned to give $\mathcal{P}$, which is monocular and hence fixed only up to a global similarity.

\paragraph{Segmentation and lifting.}
Each keyframe is segmented by a promptable open-vocabulary model \cite{sam3} with a fixed prompt set, returning scored instance masks; low-scoring masks are rejected. A mask is lifted to 3D without a separate depth step, since the front end already gives a 3D point per pixel: the mask selects the points of its own keyframe, after removing non-finite, degenerate, or low-confidence points. Masks retaining too few valid points are dropped as too marginally observed to localise.

\paragraph{Structural channel.}
Masks for the structural prompts (\emph{wall}, \emph{window}) are not treated as objects; their points accumulate across keyframes into the structural cloud $\mathcal{P}_w \subset \mathcal{P}$, the boundary from which the top-down map is formed. $\mathcal{P}_w$ is recovered without depth supervision, and its coverage is uneven (Sec.~\ref{sec:bev}).

\paragraph{Door-frame channel.}
Masks for the \emph{door frame} prompt accumulate into a third cloud, kept apart from $\mathcal{P}_w$. A door frame is not boundary during the closure; it is boundary only where visibility is measured, so it is carried separately and projected onto the grid in Sec.~\ref{sec:bev}.

\paragraph{Object channel.}
Masks for the remaining prompts yield object observations, each a voxel set, associated online into persistent tracks $\mathcal{O} = \{o_j\}$. A new observation with voxel set $A$ is matched, via a spatial index, only against tracks sharing its voxels, using the overlap coefficient
\begin{equation}
  \mathrm{IoS}(A, B) = \frac{|A \cap B|}{\min(|A|, |B|)} .
  \label{eq:ios}
\end{equation}
IoS replaces IoU so that a small fragment largely contained in a track accumulated over many views still scores high rather than being penalised by the track's size; an observation matching two tracks bridges them. Each finished track reduces to a centroid $c_j$ and an oriented box. These components are used as released; the stage is not a contribution and supplies only $\mathcal{P}_w$, the door frames, the trajectory $\mathrm{T}$, and the object tracks with their centroids and oriented boxes.

\section{The Workflow in One Diagram}
\label{app:overview}
\begin{figure*}[tp]
\centering

\tikzset{
  pcBase/.style  = {draw, rounded corners=2pt, align=center, font=\footnotesize, inner sep=3pt, line width=.5pt, minimum height=8.0mm}, pcInp/.style   = {pcBase, fill=green!12,   draw=green!55!black}, pcExist/.style = {pcBase, fill=blue!8,     draw=blue!55!black}, pcCanon/.style = {pcBase, fill=cyan!14,    draw=cyan!60!black}, pcOurs/.style  = {pcBase, fill=magenta!12, draw=magenta!70!black, font=\footnotesize\bfseries, line width=.8pt}, pcAbl/.style   = {pcBase, fill=magenta!3,  draw=magenta!45!black, dashed}, pcData/.style  = {draw, trapezium, trapezium left angle=80, trapezium right angle=100, trapezium stretches body, fill=yellow!22, draw=yellow!55!black, align=center, font=\footnotesize, inner sep=3pt, minimum height=7.0mm},
  pcDataB/.style = {pcData, align=none},
  pcAr/.style    = {-{Stealth[length=4pt,width=3pt]}, draw=black!62, line width=.6pt, rounded corners=2pt}, pcArw/.style   = {-{Stealth[length=4pt,width=3pt]}, draw=black!40, line width=.55pt, dashed, rounded corners=2pt}, pcArf/.style   = {-{Stealth[length=3.6pt,width=2.8pt]}, draw=black!35, line width=.45pt, dotted, rounded corners=2pt}, pcArd/.style = {pcAr, dashed}, pcElab/.style  = {font=\scriptsize, text=black!55, inner sep=1.5pt, fill=white, midway},
  pcGrp/.style   = {draw, dashed, rounded corners=5pt, inner sep=6pt, line width=.6pt}, pcGlab/.style  = {font=\scriptsize\itshape}, }
\newcommand{\pcs}[1]{{\scriptsize\textcolor{black!60}{#1}}} \newcommand{\pcin}[2]{\\[2pt]\includegraphics[height=#2]{#1}}
\newcommand{\pcbes}[3]{\ensuremath{%
  \vcenter{\hbox{\normalfont\shortstack{#1}}}\hspace{3pt}%
  \vcenter{\hbox{\includegraphics[height=#3]{#2}}}}}

\resizebox{\textwidth}{!}{%
\begin{tabular}{@{}c@{}}

\begin{tikzpicture}

\node[pcInp] (vid) at (1.25,-1.10) {Input video\\\pcs{monocular camera}\pcin{input_frames.png}{1.05cm}};
\node[pcExist, right=5mm of vid] (slam) {MASt3R-SLAM~\cite{mast3rslam}\pcin{mast3r_slam.png}{1.05cm}};
\node[pcData,  right=5mm of slam.east, anchor=west, yshift=.05mm] (kf) {keyframes};
\node[pcExist, right=5mm of kf]  (sam) {SAM3~\cite{sam3}\\\pcs{per-keyframe}};
\node[pcExist, right=5mm of sam] (ext) {extract point cloud\\per segmented mask};
\node[pcData, above=7mm of sam] (prm) {prompts: bag, television, laptop, remote, bed,\\
       toothbrush, table, wall, room door, window};

\draw[pcAr] (vid)  -- (slam); \draw[pcAr] (slam) -- (kf); \draw[pcAr] (kf)   -- (sam); \draw[pcAr] (prm)  -- (sam); \draw[pcAr] (sam)  -- (ext);

\node[pcData]  (traj)at (1.40,-3.30) {camera\\trajectory}; \node[anchor=north, inner sep=0] (trajI) at (1.40,-3.80) {\includegraphics[height=1.45cm]{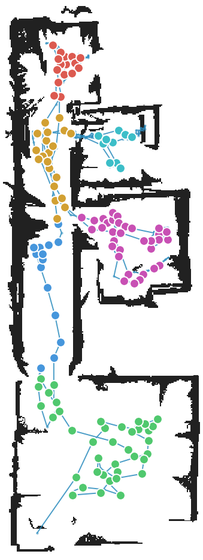}}; \node[pcData]  (pc3) at (4.30,-3.30) {3D point clouds}; \node[pcData]  (pw)  at (6.95,-3.60) {$\mathcal{P}_w$ wall\,/\,window\\point clouds\pcin{wall_cloud.png}{1.00cm}};

\node[pcData,  right=8mm of pw.east, anchor=west, yshift=3mm] (opc) {object\\point clouds};
\node[pcExist, right=4mm of opc]  (trk)  {Voxel-IoS\\tracker\\\pcs{Eq.~\eqref{eq:ios}}};
\node[pcExist, right=4mm of trk]  (cen)  {centroid $+$\\oriented box};
\node[pcData,  right=4mm of cen]  (trks) {tracks $\{o_j\}$\\extents $\{U_j\}$};

\path (slam.south) -- ++(0,-0.35) coordinate (bus);
\draw[pcAr,-] (slam.south) -- (bus);
\draw[pcAr,-] (traj.north |- bus) -- (bus) -- (pc3.north |- bus);
\draw[pcAr]   (traj.north |- bus) -- (traj.north);
\draw[pcAr]   (pc3.north  |- bus) -- (pc3.north);
\draw[pcAr] ([xshift=6mm]pc3.north) -- ++(0,0.90) -| ([xshift=4mm]ext.south west);
\draw[pcAr] (ext.south) -- ++(0,-1.10) -| node[pcElab, pos=0.24]{``wall''\,/\,``window''} (pw.north);
\draw[pcArd] (ext.south) -- ++(0,-0.75) -| node[pcElab, pos=0.55]{other prompts} (opc.north);
\draw[pcAr] (opc)  -- (trk); \draw[pcAr] (trk)  -- (cen); \draw[pcAr] (cen)  -- (trks);

\node[pcCanon] (grv) at (1.85,-7.55) {gravity $+$ Manhattan\\alignment\pcin{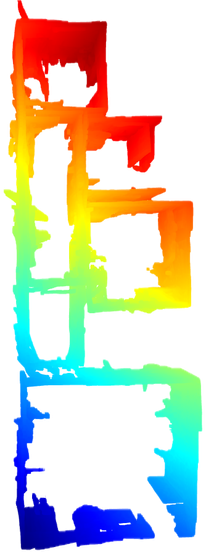}{1.75cm}}; \node[pcCanon] (prj) at (4.85,-7.55) {height cut $+$\\project to BEV};
\node[pcData]  (bev) at (7.05,-7.30) {BEV map\\$W$,\,$F$}; \node[anchor=north, inner sep=0] (bevI) at (7.05,-7.80) {\includegraphics[height=1.75cm]{bev_map.png}};

\draw[pcAr] (pw.south) -- ++(0,-1.05) -| (grv.north); \draw[pcAr] (grv) -- (prj); \draw[pcAr] (prj) -- (bev);

\node[pcOurs] (clo) at (9.95,-7.40) {Progressive\\closure\\\mdseries\pcs{Sec.~\ref{sec:rooms}}};
\node[pcData,  right=4mm of clo] (sds) {seed groups\\$\{\mathcal{S}_i\}$};
\node[pcOurs,  right=4mm of sds] (den) {ray vote\\$+$ BFS fill};
\node[pcDataB, right=4mm of den] (rms) {\pcbes{rooms\\$\{\mathcal{R}_i\}$}{room_seg.png}{1.40cm}};

\node[pcOurs] (sm) at (10.05,-9.95) {Surface majority\\\mdseries\pcs{Sec.~\ref{sec:surfmaj}}};
\node[pcAbl]  (va) at (12.80,-9.95) {visibility agreement\\\pcs{Sec.~\ref{sec:assign:vis} --- ablation}};
\node[pcData, anchor=east] (out) at ([yshift=-2.30cm]rms.east) {object-to-room\\assignments $f$\pcin{obj_assignment.png}{1.50cm}};

\draw[pcAr] (bevI.east) -| ([xshift=-4mm]clo.south); \draw[pcAr] (clo) -- (sds); \draw[pcAr] (sds) -- (den); \draw[pcAr] (den) -- (rms);

\draw[pcAr]  (rms.south) -- ++(0,-0.70) -| (sm.north);
\draw[pcArf] ([xshift=-8mm]rms.south) -- ++(0,-0.40) -| node[pcElab, pos=0.72]{fallback} (va.north east);
\draw[pcArw] (sds.south) -- ++(0,-1.45) -| (va.north west);
\draw[pcAr]  (trks.east) -- ++(0.65,0) |- node[pcElab, pos=0.30]{extents $U_j$} (out.east);
\draw[pcAr]  (sm.south) -- ++(0,-1.05) -| (out.south);
\draw[pcArw] (va.south) -- ++(0,-0.75) -| ([xshift=-6mm]out.south);

\node[inner sep=0, minimum size=0] (padL) at (10.05,-11.75) {};
\node[inner sep=0, minimum size=0] (padR) at ([yshift=-4.35cm]rms.east) {};

\begin{scope}[on background layer]
  \node[pcGrp, draw=blue!45!black, fill=blue!3, fit=(prm)(vid)(slam)(kf)(sam)(ext)(traj)(trajI)(pc3)(pw) (opc)(trk)(cen)(trks)] (gA) {}; \node[pcGrp, draw=cyan!55!black, fill=cyan!4, fit=(grv)(prj)(bev)(bevI)] (gB) {}; \node[pcGrp, draw=magenta!65!black, fill=magenta!3, fit=(clo)(sds)(den)(rms)(sm)(va)(out)(padL)(padR)] (gC) {};
\end{scope}
\node[pcGlab, text=blue!45!black, anchor=north west]
      at ([xshift=3pt]gA.south west) {};
\node[pcGlab, text=cyan!55!black, anchor=north west]
      at ([xshift=3pt]gB.south west) {canonical projection (Sec.~\ref{sec:bev})};
\node[pcGlab, text=magenta!65!black, anchor=south west]
      at ([xshift=3pt]gC.north west) {our contributions};

\end{tikzpicture}
\\[6mm]

\begin{tikzpicture}[ >=Stealth, every node/.style={font=\scriptsize}, panelborder/.style={draw=black!35, line width=0.6pt}, paneltitle/.style={font=\scriptsize\bfseries, text=black!80}, arr/.style={->, thick, black!50, line width=0.8pt}, ]

\def\panelgap{0.18}     
\def\pw{3.85}           
\def\ph{2.4}            

\pgfmathsetmacro{\xb}{\pw+\panelgap}
\pgfmathsetmacro{\xc}{2*(\pw+\panelgap)}
\pgfmathsetmacro{\xd}{3*(\pw+\panelgap)}
\pgfmathsetmacro{\xe}{4*(\pw+\panelgap)}

\node[panelborder, minimum width=\pw cm, minimum height=\ph cm, anchor=north west] (pa) at (0,0) {\includegraphics[width=\pw cm, height=\ph cm, keepaspectratio=false]%
     {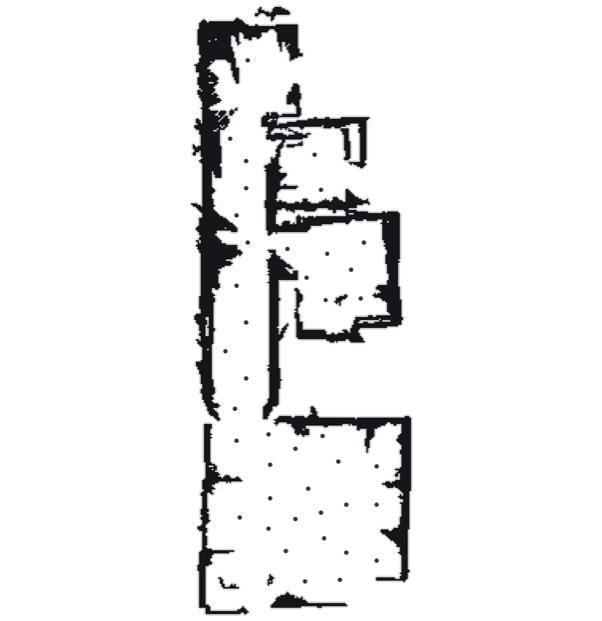}}; \node[paneltitle, anchor=south, yshift=2pt] at (pa.north) {(a)~$t=1$: seeds placed}; \node[font=\tiny\itshape, text=black!50, anchor=north, yshift=-2pt] at (pa.south) {all 48 seeds active};

\node[panelborder, minimum width=\pw cm, minimum height=\ph cm, anchor=north west] (pb) at (\xb cm, 0) {\includegraphics[width=\pw cm, height=\ph cm, keepaspectratio=false]%
     {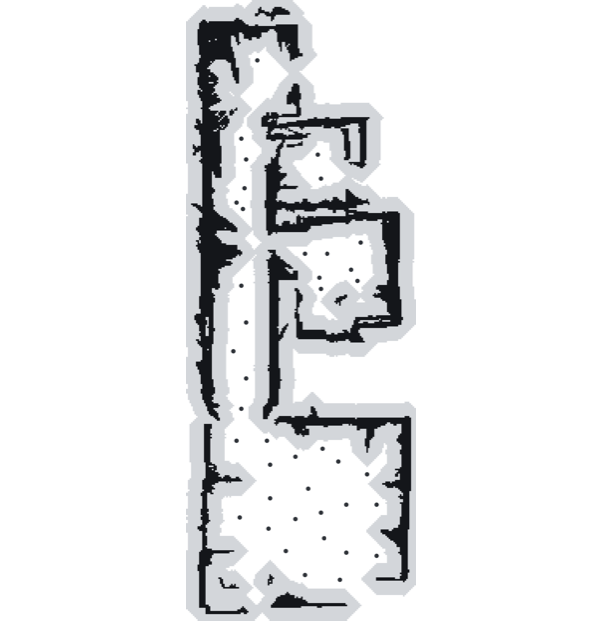}}; \node[paneltitle, anchor=south, yshift=2pt] at (pb.north) {(b)~$t=9$: walls thicken}; \node[font=\tiny\itshape, text=black!50, anchor=north, yshift=-2pt] at (pb.south) {narrow gaps closing};

\node[panelborder, minimum width=\pw cm, minimum height=\ph cm, anchor=north west] (pc) at (\xc cm, 0) {\includegraphics[width=\pw cm, height=\ph cm, keepaspectratio=false]%
     {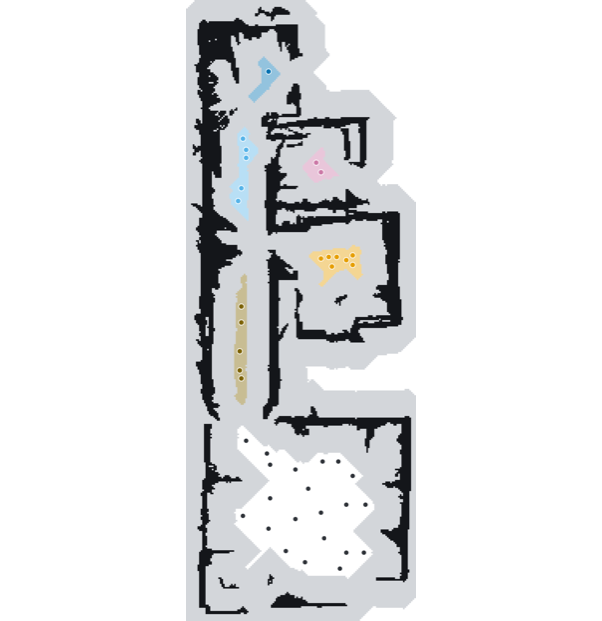}}; \node[paneltitle, anchor=south, yshift=2pt] at (pc.north) {(c)~$t=17$: sealing}; \node[font=\tiny\itshape, text=black!50, anchor=north, yshift=-2pt] at (pc.south) {19/48 frozen at own scale};

\node[panelborder, minimum width=\pw cm, minimum height=\ph cm, anchor=north west] (pd) at (\xd cm, 0) {\includegraphics[width=\pw cm, height=\ph cm, keepaspectratio=false]%
     {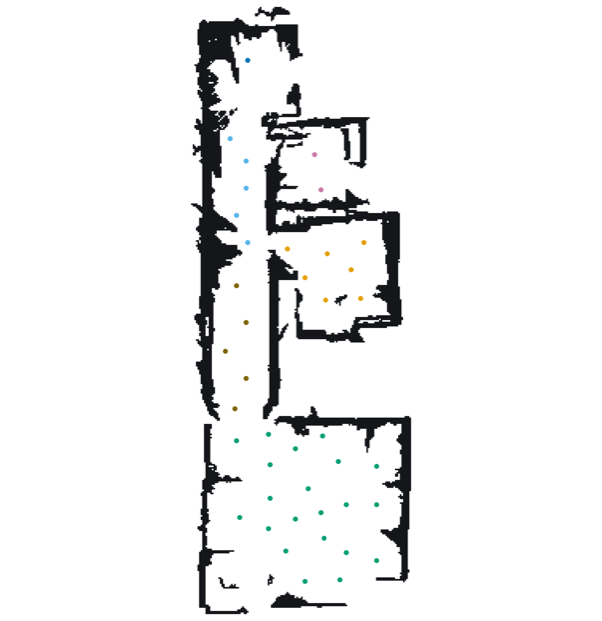}}; \node[paneltitle, anchor=south, yshift=2pt] at (pd.north) {(d) Rooms $\{\mathcal{R}_i\}$}; \node[font=\tiny\itshape, text=black!50, anchor=north, yshift=-2pt] at (pd.south) {5 rooms, seeds at $p^{\mathrm{init}}$};

\node[panelborder, minimum width=\pw cm, minimum height=\ph cm, anchor=north west] (pe) at (\xe cm, 0) {\includegraphics[width=\pw cm, height=\ph cm, keepaspectratio=false]%
     {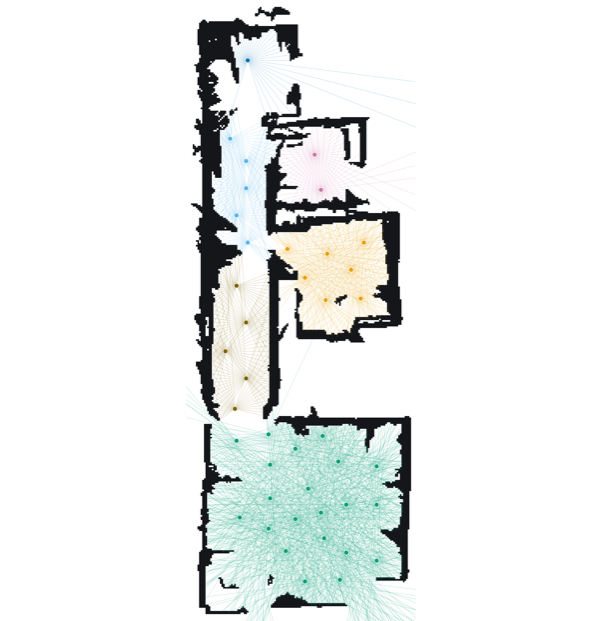}}; \node[paneltitle, anchor=south, yshift=2pt] at (pe.north) {(e) Sightlines}; \node[font=\tiny\itshape, text=black!50, anchor=north, yshift=-2pt] at (pe.south) {rays pruned across rooms};

\foreach \from/\to in {pa/pb, pb/pc, pc/pd}{ \draw[arr] ([xshift=1mm]\from.east) -- ([xshift=-1mm]\to.west); }

\draw[arr, dashed] ([xshift=1mm]pd.east) -- ([xshift=-1mm]pe.west);

\draw[thick, blue!50!black!60] ([yshift=-0.55cm]pa.south west) -- ([yshift=-0.55cm]pd.south east) node[midway, below=6pt, align=center, font=\scriptsize, text=blue!60!black] {progressive closure — each room seals at its own scale (Eq.~\eqref{eq:freeze})};

\end{tikzpicture}

\end{tabular}%
}

\caption{The full workflow in one diagram. A monocular RGB video is processed by MASt3R-SLAM~\cite{mast3rslam}, which returns keyframes, a dense point cloud and the camera trajectory. SAM3~\cite{sam3} segments each keyframe against a fixed prompt set; masks for \emph{wall} and \emph{window} accumulate into the structural cloud $\mathcal{P}_w$, and the remaining prompts become object observations tracked by voxel-IoS matching. The structural cloud is aligned, height-cut and rasterised into a binary top-down map $(W,F)$. Our contributions are shown in pink. \emph{Progressive closure} grows the boundaries of $W$ until every pocket saturates, seeded by the projected camera trajectory rather than by sampling the map (Sec.~\ref{sec:rooms}); its output is a set of seed groups $\{\mathcal{S}_i\}$, one per recovered room. The retained sightlines of those groups then vote for a label at each free cell, and a breadth-first fill restricted to free space completes the labelling, giving the dense rooms $\{\mathcal{R}_i\}$. The two assignment rules consume different products of this chain, which is why the flow branches at $\{\mathcal{S}_i\}$. \emph{Surface majority} needs the dense labels: it assigns each track to the room holding most of its reconstructed extent $U_j$ (Sec.~\ref{sec:surfmaj}). The dashed box is the \emph{visibility-agreement} rule it is compared against in Sec.~\ref{sec:res-ablation}, which scores a track by the sightlines it shares with each room's seeds (Sec.~\ref{sec:assign:vis}) and therefore requires no dense labelling at all; it consults the labels only through the dotted fallback edge, taken when no room reaches a nonzero agreement.}
\label{fig:overview-full}
\end{figure*}

Figs.~\ref{fig:pipeline-data} and~\ref{fig:pipeline} of the main paper split the pipeline in two so that each half fits a column. Fig.~\ref{fig:overview-full} draws the same workflow in one piece, with every intermediate product and the closure sequence in a single view.

\section{Properties of the Closure}
\label{app:properties}

\emph{The room count is emergent}: it starts as the number of components of $\Phi$ and is set by geometry, not supplied; the corrections of Sec.~\ref{sec:cleanup} may then discard a region or found one. \emph{Each room is resolved at its own scale}: a small room behind narrow apertures and a large one behind a wide opening both close, at different iterations, which one radius cannot achieve. \emph{Apertures are not classified}: the closure treats a doorway and an unobserved gap identically, needing only that an opening be narrow, which is the property that suits it to monocular reconstruction. Detected door frames enter only where visibility is measured, never as a partition criterion.

\section{Rooms and Cleanup}
\label{app:cleanup}

\paragraph{Regions that never sealed.}
A pocket can stop growing because it ran out of admissible cells, or because it ran into the edge of the raster. Only the first is a room; the second is capture lying outside the building, bounded by where the reconstruction stops rather than by a wall. The two are told apart by the building footprint $\Omega_{\mathrm{fp}}$ of Sec.~\ref{sec:bev}, the filled outer contour of the projected full cloud, which marks the area the reconstruction covers. A region is discarded when
\begin{equation}
  \frac{\bigl| \mathcal{R}_i \setminus \Omega_{\mathrm{fp}} \bigr|}
       {\bigl| \mathcal{R}_i \bigr|} > \phi ,
  \label{eq:unsealed}
\end{equation}
with $\phi$ a fixed fraction. Its seeds are kept and passed to the rule below.

\paragraph{Seeds that cannot see their region.}
A seed relocating under Eq.~\eqref{eq:relocate} can end up in a neighbouring space, where it witnesses the wrong region. We score each seed against the centroid $c_i$ of every region with the agreement measure $a(\cdot,\cdot)$ of Sec.~\ref{sec:assign:vis}. Seeds are visited in ascending order of $a(p, c_{i(p)})$, their agreement with their own region, so the worst-placed seed is resolved first; seeds released by the rule above have no region of their own and are visited before any of them. A seed whose own agreement reaches $\tau$ stays. Otherwise it moves to the region of highest agreement if that exceeds $\tau$; if none does, the seed founds a region of its own. A region founded this way is immediately a candidate for the seeds still to be visited, so a seed belonging nowhere can gather the neighbours that belong with it.

\paragraph{Doorways as boundary.}
The third correction is stated in Sec.~\ref{sec:cleanup}: every visibility test is computed on $W \cup \Delta$ rather than on $W$, so a fan cannot cross an open doorway. A seed whose own position falls on $\Delta$ sees nothing in any direction and is removed rather than re-homed, since it stands in the aperture and belongs to neither adjoining region.

\paragraph{Displaced seeds.}
A seed may slip through an aperture into an adjacent space before it seals, ending in the wrong region. Within each room of at least $n_{\min}$ seeds we discard those whose displacement $\lVert p^{\mathrm{init}} - p^{\mathrm{final}} \rVert$ exceeds $\bar{d} + k\,s_d$ (mean and s.d.\ over the room's seeds); smaller rooms are left untouched.

\paragraph{Room representation.}
Freezing fixes how many rooms there are and which seeds witness each, at every room's own scale; it does not by itself fix their extent, since the frozen pocket is the region as of its seal and is eroded by the dilation that sealed it. A separate delineation pass (Sec.~\ref{sec:delineate}) recovers full extents, and those extents are what the room evaluation of Sec.~\ref{sec:setup} scores. Every seed is scored and casts its sightlines from its sampled position $p^{\mathrm{init}}$, not from wherever relocation left it.

\section{Object Placement Algorithm}
\label{app:assign-alg}
Algorithm~\ref{alg:assign} states the visibility-agreement procedure of Sec.~\ref{sec:assign} in full.

\begin{algorithm}[!htbp]
\caption{Object placement by visibility agreement}
\label{alg:assign}
\begin{algorithmic}[1]
\Require tracks $\mathcal{O}$ with centroids $\{c_j\}$; seed clusters $\{\mathcal{S}_i\}$; occupied set $W$; door frames $\Delta$; grid $\Omega$; projection $\pi_Q$; parameters $K, \ell_s, \ell_o, \tau_a$
\Ensure  assignment $f$, route flag $g$
\ForAll{$q \in \bigcup_i \mathcal{S}_i$}
  \State $R^{s}(q) \gets$ cast $K$ rays from $q$ on $W \cup \Delta$; stop at $W \cup \Delta$, $\partial\Omega$, or $\ell_s$ steps
\EndFor
\ForAll{$o_j \in \mathcal{O}$}
  \State $p \gets \pi_Q(c_j)$
  \State $R^{o}(p) \gets$ cast $K$ rays from $p$ on $W \cup \Delta$; stop at $W \cup \Delta$, $\partial\Omega$, or $\ell_o$ steps
  \ForAll{rooms $i$}
    \State $s_i \gets \frac{1}{|\mathcal{S}_i|} \sum_{q \in \mathcal{S}_i} a(p, q)$ \Comment{Eqs.~\eqref{eq:agreement},~\eqref{eq:objscore}}
  \EndFor
  \State $i^{\star} \gets \arg\max_i s_i$
  \If{$s_{i^{\star}} \ge \tau_a$}
    \State $f(o_j) \gets i^{\star}$;\quad $g(o_j) \gets \textsc{vis}$
  \Else
    \State $q^{\star} \gets \arg\min_{q \in \bigcup_i \mathcal{S}_i} \lVert p - q \rVert_1$
    \State $f(o_j) \gets$ room of $q^{\star}$;\quad $g(o_j) \gets \textsc{fallback}$
  \EndIf
\EndFor
\State \Return $f, g$
\end{algorithmic}
\end{algorithm}

\section{Evaluation Protocol in Full}
\label{app:metrics}
The first is the \emph{overlap} criterion established for this task~\cite{bormann2016} and adopted throughout the scene-graph literature~\cite{hughes2022hydra,werby2024hovsg}. Writing $\mathcal{R}_e$ for the predicted rooms and $\mathcal{R}_g$ for the ground-truth regions,
\begin{equation}
\begin{aligned}
  P_{\mathrm{ov}} &= \frac{1}{|\mathcal{R}_e|}\sum_{r_e\in\mathcal{R}_e}
    \max_{r_g\in\mathcal{R}_g}\frac{|r_e\cap r_g|}{|r_e|}, \\[2pt]
  R_{\mathrm{ov}} &= \frac{1}{|\mathcal{R}_g|}\sum_{r_g\in\mathcal{R}_g}
    \max_{r_e\in\mathcal{R}_e}\frac{|r_e\cap r_g|}{|r_g|}.
\end{aligned}
\label{eq:pr}
\end{equation}
The maximum is taken independently in each direction, so the pairing is neither one-to-one nor thresholded. Low $P_{\mathrm{ov}}$ indicates under-segmentation and low $R_{\mathrm{ov}}$ over-segmentation.

The second is an \emph{instance} criterion: predictions and regions are matched one-to-one by Hungarian assignment on IoU, a pair counts as a true positive when its IoU exceeds $\tau \in \{0.25, 0.5\}$, and we report $P_\tau = \mathrm{TP}/|\mathcal{R}_e|$, $R_\tau = \mathrm{TP}/|\mathcal{R}_g|$, $F_{1,\tau}$, and mean IoU over matched pairs. It is the stricter of the two, as it penalises a prediction that overlaps the right region loosely, which Eq.~\eqref{eq:pr} rewards.

\paragraph{Scoring domain.} Both criteria are computed on the cells of the top-down grid, the plane in which the partition is defined, whereas prior work computes them over 3D voxels or surface points. The difference is not cosmetic. Sampling the annotated mesh weights a room by the surface it contains rather than the floor it covers, so a cluttered bedroom carries several times the evidence of an empty hall of equal size; and $16.9\%$ of the sampled surface carries no region identifier at all, mostly ceilings and upper walls. Scoring on the grid removes both effects. Surface-point figures appear alongside in Table~\ref{tab:published} for comparability, and we state the domain wherever our numbers sit beside published ones. Scores are averaged over the ten floors, so each floor weighs equally and the means describe the same quantity as the tests below.

\section{Three Findings Across the Results}
\label{app:findings}
Three findings carry across Secs.~\ref{sec:res-quant}--\ref{sec:res-fail}. The first is that the room-count gap drives most of Table~\ref{tab:rooms-main}: our reimplementation of HOV-SG predicts 44 rooms for 72 observed regions where we predict 74, so its conservatism buys precision at the $0.25$ threshold ($0.98$) at the cost of recall ($0.60$), while ours trades a little high-threshold precision for a partition close to the true count and a better $F_1$ at both thresholds. The second is that the doorway seal is the one stage whose contribution is separable at this sample size, improving all ten floors for $+0.034$ mIoU; it is also the one stage that consumes an input the baseline does not receive, which Sec.~\ref{sec:baselines} states. The third is that the failures of Sec.~\ref{sec:res-fail} are structural rather than incidental, and the three floors discussed there are specific evidence for them rather than a generic caveat.

\section{Phase A/B Walkthrough}
\label{app:walkthroughfig}

\begin{figure}[!t]
\centering
\setlength{\tabcolsep}{1.5pt}
\renewcommand{\arraystretch}{0.78}
\newcommand{\qualcell}[1]{\includegraphics[width=0.238\columnwidth]{#1}}
\newcommand{\qualblank}{\rule{0pt}{0pt}}
\begin{tabular}{@{}cccc@{}}
\qualcell{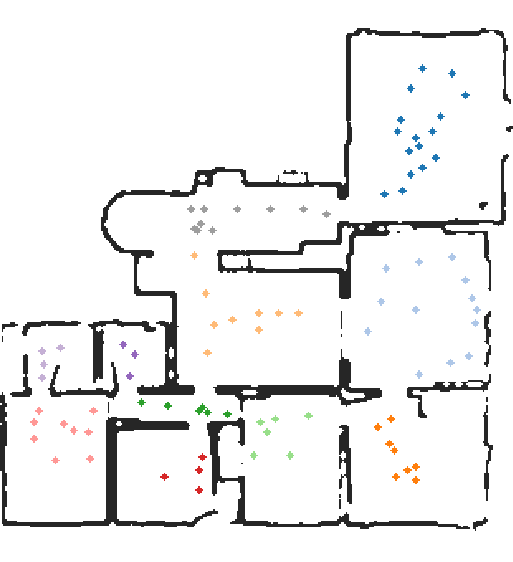} &
\qualcell{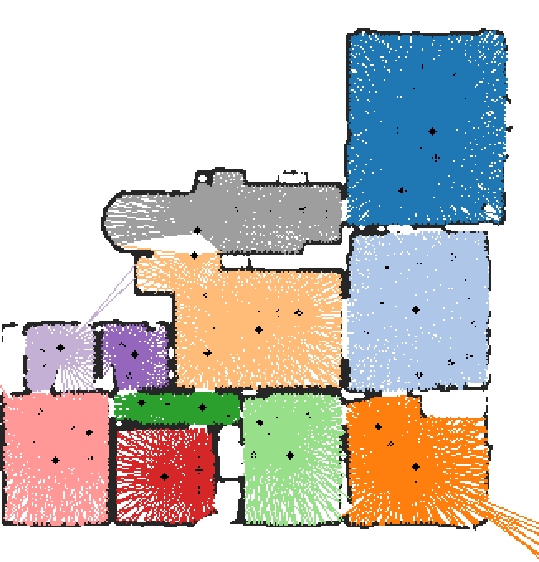} &
\qualcell{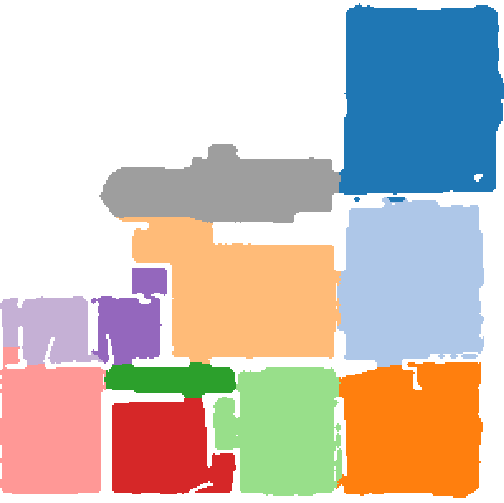} &
\qualcell{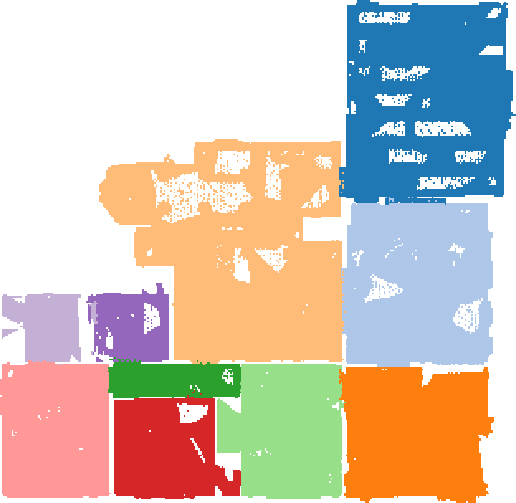} \\
{\scriptsize (1)} & {\scriptsize (2)} & {\scriptsize (3)} & {\scriptsize (4)} \\
\end{tabular}

\caption{\textbf{Phase A/B walkthrough, \texttt{00824-Dd4bFSTQ8gi} floor 0, final-method (camera-pose) seeding.} 1: 83 seeds (of 85 recorded poses on this floor; the other 2 do not survive de-duplication), coloured by eventual cluster. 2: kept sightlines after cross-cluster pruning (Sec.~\ref{sec:assign}). 3--4: delineated rooms vs.\ matched GT regions, same colouring. Nine of 11 predicted rooms match a GT region at both thresholds (P 0.82, R 0.90, $F_1$ 0.86); mean IoU over matched pairs is 0.695, with one unmatched room.} \Description{Four top-down panels in a row showing the seeds coloured by cluster, the retained sightlines, the delineated rooms and the ground-truth regions for one floor.}
\label{fig:pipeline_walkthrough}
\end{figure}

Figure~\ref{fig:pipeline_walkthrough} follows one floor through the two
phases of Sec.~\ref{sec:rooms}, from the seeds the trajectory supplies to the
delineated rooms scored against the annotated regions.

\section{Room-Partitioning Families}
\label{app:rwrooms}
Clearance-based methods locate rooms where the distance to the nearest obstacle is large. HOV-SG~\cite{werby2024hovsg} thresholds this field, takes the surviving components as room cores, and floods outward from them; Hydra~\cite{hughes2022hydra} maintains a graph of locations lying at safe distance from obstacles and partitions that graph into rooms; Point2Graph~\cite{point2graph2025} detects regions from a density map with enhanced borders. All depend on clearance contracting at every passage between one room and the next.

Connectivity-based methods instead reduce free space to a skeleton or adjacency graph and sever it where it narrows~\cite{thrun1998learning, friedman2007vrf}; ROSE2~\cite{rose2_2022} first recovers dominant wall directions so that clutter does not produce spurious constrictions, and OccuSG~\cite{occusg2026} decomposes occupancy directly at narrow passages. Learned methods take a different route, regressing room polygons from a map or point cloud~\cite{liu2018floornet, chen2019floorsp, chen2022heat, yue2023roomformer, frinet2024}, at the cost of annotated training data and, in most cases, layouts close to rectilinear.

\section{Canonical Frame in Full}
\label{app:canon}
\paragraph{Gravity alignment.}
Planar surfaces are extracted from $\mathcal{P}$ by sequential RANSAC, each plane removed before the next is fitted. Among candidates whose normal is approximately vertical, the one with most inliers is taken as the floor, and $R_g \in SO(3)$ maps its normal onto the vertical axis.

\paragraph{Manhattan correction.}
A rotation about the vertical remains. Points in a central height band with near-horizontal normals are treated as wall samples, their normals projected to the horizontal plane, reduced to orientations $\theta = \operatorname{atan2}(n_y, n_x)$, folded modulo $\pi/2$ and histogrammed. The modal orientation gives a yaw $R_y$ that aligns the dominant wall direction with the grid, so $Q = R_y R_g$. This reduces aliasing of wall lines only; the closure of Sec.~\ref{sec:rooms} is isotropic and requires no Manhattan layout.

\paragraph{Height selection.}
A boundary matters only where an agent can meet it, and the reconstruction is sparsest near the ceiling. We cut the transformed cloud at the mid-height of its own vertical extent and rasterise the half below, which spans the height envelope such an agent occupies. Expressed as a fraction of the scene's own height, the cut requires no metric scale.

\section{Further Limitations}
\label{app:failmodes}
\paragraph{Junctions with no constriction.}
The second mode is not a wide opening but the absence of any opening to find. On \texttt{00829\-QaLdnwvtxbs} floor~0 the walkway and the room adjoining it share a frontier of 262 free cells, against 4 cells for a doorway elsewhere on the same floor and a median of 39 over all adjacent pairs. The two spaces meet along an entire wall line, with no door, no frame and no wall stub, so the growth has no width minimum to seal and the pocket spanning both never divides. The walkway seeds relocate into the room as the boundary thickens and the room is recovered as part of the walkway, at IoU $0.24$ against $0.48$ to $0.85$ for every other region on that floor. Unlike the first mode this is not a matter of relative extent: the room is not shallow, and the architecture simply provides no minimum for a width-based rule to act on. Recovering such a boundary requires evidence the top-down map does not carry, such as a change in ceiling height or floor material.

\paragraph{Boundaries placed short.}
\texttt{00877-4ok3usBNeis} floor~1 is one of three floors on which HOV-SG scores higher ($0.755$ against $0.689$). The room count is not the problem: we recover 8 rooms for 9 observed regions where HOV-SG recovers 3. Several of our boundaries fall short of the annotated ones, so rooms that are correctly separated overlap their regions loosely, and mIoU rewards a smaller number of well-fitted rooms over a larger number of approximate ones. It is also the floor on which visibility agreement most clearly beats surface majority (ARI $0.641$ against $0.513$, Sec.~\ref{sec:res-ablation}).

\section{Seed Relocation in Full}
\label{app:relocate}
\paragraph{Relocation.}
Thickening the boundary encroaches on seeds. A seed whose region is not yet declared and whose clearance has fallen below $\kappa$ moves to the nearest cell that is still admissible,
\begin{equation}
\begin{aligned}
  p' &= \arg\min_{q \,\in\, A^{(t)}(p)} d_g(p, q), \\[2pt]
  A^{(t)}(p) &= \left\{ q :
  \begin{aligned}
    &D^{(t)}(q) \ge \kappa, \\
    &\lVert q - p_u \rVert \ge \sigma \ \ \forall u \neq p, \\
    &d_g(p, q) \le \mu
  \end{aligned}
  \right\},
\end{aligned}
  \label{eq:relocate}
\end{equation}
where $d_g$ is geodesic distance within $F^{(t)} \setminus \Phi^{(t)}$ and $\mu$ caps the displacement. If no admissible cell is reachable the seed stays where it is. Two details matter. Distance is measured along free space rather than in straight line, so a seed cannot step across a wall to reach open space on the far side. Declared regions are excluded from the search, so a seed cannot drift into a room that has already been resolved.

\section{Progressive Closure in Full}
\label{app:closure-detail}

Each step dilates the occupied set by $B_\rho$ outside the declared cells and leaves the declared cells untouched:
\begin{equation}
\begin{aligned}
  W^{(t)} &= \bigl( \delta_{B_\rho}\!\bigl(W^{(t-1)}\bigr) \setminus \Phi^{(t-1)}
  \bigr) \cup \bigl( W^{(t-1)} \cap \Phi^{(t-1)} \bigr), \\
  F^{(t)} &= \Omega \setminus W^{(t)} .
\end{aligned}
  \label{eq:dilate}
\end{equation}
A cell qualifies when it is at least $\kappa$ from the boundary and at least $\sigma$ from every current seed:
\begin{equation}
  V^{(t)} = \bigl\{\, u \in F^{(t)} \setminus \Phi^{(t-1)} :
            D^{(t)}(u) \ge \kappa,\ d_{\mathcal{S}}^{(t)}(u) \ge \sigma \,\bigr\},
  \label{eq:admissible}
\end{equation}
with $d_{\mathcal{S}}^{(t)}(u) = \min_{p \in \mathcal{S}^{(t)}} \lVert u - p \rVert_2$. A pocket is saturated when it holds no qualifying cell; every saturated pocket containing a seed is declared:
\begin{equation}
  \Phi^{(t)} = \Phi^{(t-1)} \cup \bigcup_{C \in \mathcal{C}^{(t)}_{\mathrm{new}}} C ,
  \label{eq:freeze}
\end{equation}
where $\mathcal{C}^{(t)}_{\mathrm{new}} = \{\, C \in \mathcal{C}^{(t)} : C \cap \mathcal{S}^{(t)} \neq \emptyset,\ C \cap V^{(t)} = \emptyset \,\}$ and $\mathcal{C}^{(t)}$ are the components of $F^{(t)} \setminus \Phi^{(t-1)}$.

\section{Reading Table 1}
\label{app:reading}
\paragraph{Reading the table.}
HOV-SG's construction is conservative. In Table~\ref{tab:rooms-main} it predicts 44 rooms for 72 observed regions and reaches $P@.25 = 0.98$ by rarely predicting a room that overlaps nothing, at the cost of $R@.25 = 0.60$, which means two of every five observed regions have no matching prediction. Precision alone therefore rewards predicting fewer rooms and must be read against $n_{\mathrm{pred}}$. Ours predicts close to the true count, 74 against 72, and is the more balanced partition: it gains $0.31$ recall at the $0.25$ threshold for $0.10$ precision, and leads on $F_1$ at both thresholds. The stricter threshold shows where it is weaker, precision falling to $0.81$ against HOV-SG's $0.91$, so our boundaries are more often approximately right than exactly right.

\section{Ground-Truth Provenance}
\label{app:gtprov}
Ground truth requires no annotation on our part. Room extents come from projecting HM3D-Semantics region labels onto the same grid as our top-down map, and object-to-room membership is read from instance annotations that already carry their containing region. Room scores are computed over the regions the trajectory actually observed, since a region with negligible observed extent cannot be recovered by any method; this leaves \textbf{72} of the 105. Object scores use all 105, because an object's assignment is defined whether or not its region was well covered.

\section{Recurring Failure Modes}
\label{app:failuremodes}
(1)~\emph{Wide openings.} A space joined to its neighbour by an opening comparable to its own extent does not enclose before the neighbour does, and the two are recovered as one. (2)~\emph{Junctions with no constriction.} Where two spaces meet flush, the growth has no width minimum to seal and they never divide. (3)~\emph{Unobserved regions.} Where the capture leaves a region largely unseen, the boundary is too sparse to enclose and no room is declared. Of the 105 annotated regions across the ten floors, 33 fall below the observation threshold of Sec.~\ref{sec:metrics} and are excluded from the room evaluation. (4)~\emph{Alignment failure.} Where the recovered vertical is wrong, the top-down map is not a floor plan and the partition is meaningless. (5)~\emph{Track fragmentation.} A single physical object recovered as several tracks is assigned several times, which inflates the object count and can split one object across rooms.

\section{Ablations in Full}
\label{app:ablation-detail}
\paragraph{Seed placement.}
Seeding from the camera trajectory gives mIoU $0.802$ against $0.781$ for farthest-point sampling, and is ahead on nine of the ten floors, the single exception losing by $0.064$. The margin does not reach significance ($p = 0.064$) and we report it as a trend. Its source is the mechanism described in Sec.~\ref{sec:seeds}, in that a trajectory records where the agent stood and therefore places seeds inside the rooms the walk entered rather than across an eligible pool. The procedural advantage is clearer: camera-pose seeding requires neither an eligibility pool nor a random selection step, since a recorded pose is free space by construction, and no random choice is made, so the segmentation is deterministic and admits no variance over seeds.

\paragraph{Doorway seal.}
Treating detected door frames as boundary wherever visibility is measured raises mIoU from $0.768$ to $0.802$ and improves all ten floors, giving $p = 0.002$, the smallest value the test can return for ten paired samples. The recovered room count moves from 71 to 74 against 72 observed regions. This is the only ablation reported here that separates its variants, and also the only one that relies on an input the baseline does not receive (Sec.~\ref{sec:baselines}).

\paragraph{Object-assignment rule.}
Surface majority gives ARI $0.696$ against $0.689$ for visibility agreement, at an identical room count since only the assignment stage differs. It leads on seven of ten floors, but $p = 0.193$, and on \texttt{00877} floor~1 visibility agreement is better by $0.128$, which outweighs several smaller differences in the other direction. We adopt surface majority because it leads on most floors, because it requires no visibility threshold, and because it is considerably cheaper, the ray rule costing $n_{\text{obj}} n_{\text{seed}} n_{\text{ray}}^{2}$ segment tests per floor. We do not adopt it on the grounds that the two rules are separable at this sample size.

\section{The Three Corrections}
\label{app:corrections}
Closure alone leaves three characteristic errors, and we correct each. A region bounded by the edge of the raster rather than by a wall never satisfies the enclosure criterion, and is discarded. A seed that cannot see the region it was assigned to is re-homed to the region it can see. Detected door frames are treated as boundary wherever visibility is measured, which prevents sightlines from crossing an open doorway; these are an additional input, supplied by SAM3 \cite{sam3} in the monocular pipeline and by the dataset annotations on HM3D, and Sec.~\ref{sec:res-ablation} reports what they contribute.

\section{Implementation Details}
\label{sec:impl}
\label{app:impl}

\paragraph{Reconstruction and segmentation.} MASt3R-SLAM runs uncalibrated, single-threaded. SAM3 is queried per keyframe with a fixed 11 label prompt set: \emph{wall} and \emph{window} are the structural prompts of Sec.~\ref{sec:recon} ($\mathcal{P}_w$); the remaining eight (\emph{bag}, \emph{television}, \emph{door frame}, \emph{remote}, \emph{table}, \emph{bed}, \emph{laptop}, \emph{toothbrush}, \emph{wall}, \emph{window}, \emph{room door}) are tracked as objects. Detections below score $0.8$ are rejected, and points with confidence below $1.5$ or non-finite/degenerate coordinates are discarded before lifting. Object voxels use $0.02\,\mathrm{m}$ cells (the same constant used for both the tracker and the visualisation-only global voxel map); the global point cloud $\mathcal{P}$ is voxel-downsampled at $0.05\,\mathrm{m}$ with statistical outlier removal.

\paragraph{Canonical projection.} The wall band for Manhattan alignment is the $[0.25, 0.75]$ height-percentile slab; the BEV canvas is $1024\times1024$ at $0.05\,\mathrm{m}$/px, subsequently cropped to $\Omega_{\mathrm{obs}}$, whose extent varies by scene (Table~\ref{tab:scenewise} reflects this — no two floors occupy the same crop size). Denoising uses a $3\times3$ morphological opening and drops connected components under 50 px. \paragraph{Progressive closure (Sec.~\ref{sec:rooms}).} \(\rho{=}1\) px/iteration, \(\kappa{=}5\) px, \(\sigma{=}5\) px, \(T{=}600\), relocation cardinal test \(c_m{=}4\), relocation cap \(\mu{=}400\) (Dijkstra steps), outlier cutoff \(k{=}2.0\) applied to clusters of \(n_{\min}{\ge}4\) seeds. These five (\(\rho,\kappa,\sigma,\mu,k\)) are shared by all three seeding variants below and were not re-tuned per variant.

\paragraph{Seeding (Sec.~\ref{sec:seeds}, ablated in Sec.~\ref{sec:res-ablation}).} FPS draws up to \(n_s{=}100\) seeds from an eligible pool capped at 20\,000 cells with placement cardinal test \(c_p{=}4\), first point fixed by RNG seed 42 for reproducibility. Camera-pose seeding takes every recorded frame pose on the floor (height tolerance $0.5\,\mathrm{m}$, no subsampling — stride 1) as a candidate seed directly, with no pool or cardinal test, since a frame position is free space by construction. Uniform (random) sampling of the same eligible pool as FPS is implemented (\texttt{place\_candidates} in \texttt{room\_detection.py}) but retained only as commented-out code; it produced the closure-sequence panels of Fig.~\ref{fig:pipeline} and was not run under the current evaluation harness.

\paragraph{Delineation and sightlines (Sec.~\ref{sec:delineate}--\ref{sec:assign}).} \(K{=}50\) rays/seed, seed sightline cap \(\ell_s{=}150\) px; delineated regions under 40 px are dropped, and rooms holding fewer than 2000 annotated 3D points are dropped at the point-labelling stage; the interior fill is confined to the footprint eroded by 6 px.

\paragraph{Object-to-room assignment.} The two variants compared in Sec.~\ref{sec:res-ablation} (\emph{object surface majority} and \emph{object-seed visibility voting}) were produced by a scoring harness that is not part of this code snapshot — only its outputs (\texttt{object\_eval/}) are available. We therefore report its results in Sec.~\ref{sec:res-quant}--\ref{sec:res-ablation} without restating its internal thresholds; \emph{object-seed visibility voting} is, by its stored description, the ray-agreement rule of Eq.~\eqref{eq:agreement}--\eqref{eq:objscore}.


\section{Systems that Recover a Room Layer}
\label{app:positioning}
Table~\ref{tab:positioning} groups the systems that maintain a room layer by the input each requires.


\begin{table}[!htbp]
\caption{Systems that recover a room layer, by the input they require.
\cmark{} = operates without that input. Only the last two rows are free of
poses, depth and calibration; of those, ours is the only one whose room layer is
recovered geometrically, without a learned room-transition prior.
$^{\ast}$Mono-Hydra is pose-free in the camera sense but requires IMU.}
\label{tab:positioning}
\small
\centering
\setlength{\aboverulesep}{0pt}\setlength{\belowrulesep}{0pt}
\setlength{\extrarowheight}{0.4ex}
\begin{fittable}
    \begin{tabular}{lcccl}
\toprule
System & RGB only & Pose-free & Calib-free & Room layer from \\
\midrule
Kimera~\cite{rosinol2021kimera}      & \xmark & \xmark & \xmark & free-space graph \\
Hydra~\cite{hughes2022hydra}         & \xmark & \xmark & \xmark & places graph, cut at constrictions \\
HOV-SG~\cite{werby2024hovsg}         & \xmark & \xmark & \xmark & clearance threshold + flooding \\
Clio~\cite{maggio2024clio}           & \xmark & \xmark & \xmark & task-driven clustering \\
OccuSG~\cite{occusg2026}             & \xmark & \xmark & \xmark & occupancy decomposition \\
Point2Graph~\cite{point2graph2025}   & \multicolumn{3}{c}{\emph{point cloud input}} & border + region detection \\
Mono-Hydra~\cite{monohydra}          & \cmark & \xmark$^{\ast}$ & \xmark & free-space graph \\
LEXI-SG~\cite{lexisg2026}            & \cmark & \cmark & \cmark & learned room-transition prior \\
\addlinespace
\ourrow Ours                         & \cmark & \cmark & \cmark & progressive boundary closure \\
\bottomrule
\end{tabular}
\end{fittable}
\end{table}

\section{Published Numbers in Context}
\label{app:published}

\paragraph{Why published numbers are not a comparison.}
The closest prior systems report room quality on overlapping HM3D scenes, and Table~\ref{tab:published} places those figures beside ours for context. They are not a controlled comparison and we do not treat them as one: Hydra and HOV-SG consume posed RGB-D, LEXI-SG's stronger variant consumes ground-truth poses and simulator depth, the published scores are computed over 3D free-space voxels or floor-plane points rather than our grid cells, and they are averaged per scene where we average per floor. Our controlled comparison is the one in Table~\ref{tab:rooms-main}, where every method consumes the identical top-down map produced by our own reconstruction  and is scored by the identical harness. LEXI-SG is the nearest work by input assumption --- monocular RGB without poses or calibration --- and we could not run it directly: its code was public at the time of submission.


\label{app:scenewise}

\section{Scene-Wise Analysis}
\label{sec:res-scenewise}

\begin{figure}[!htbp]
\centering
\includegraphics[width=\linewidth]{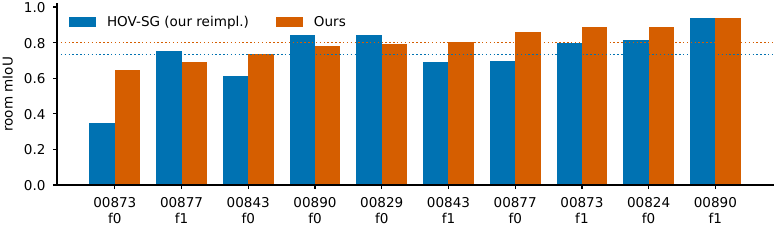}
\caption{Table~\ref{tab:scenewise} plotted per floor, sorted by our mIoU. Dotted lines mark the two
methods' mean-over-floors totals from Table~\ref{tab:rooms-main} (0.734 / 0.802). Per-floor mIoU
varies considerably for both methods (HOV-SG: 0.350--0.936; ours: 0.647--0.939), so neither total is
representative of every floor.}
\label{fig:scene-results}
\end{figure}

Table~\ref{tab:scenewise} and Figure~\ref{fig:scene-results} break Table~\ref{tab:rooms-main}'s room-mIoU row down by floor. The margin over HOV-SG is not uniform: it is largest on \texttt{00873} floor~0 (0.647 vs.\ 0.350 — HOV-SG collapses an 11-region floor into 2 predicted rooms there, Sec.~\ref{sec:res-fail}) and reversed on \texttt{00829} floor~0, \texttt{00877} floor~1 and \texttt{00890} floor~0, where HOV-SG's more conservative room count happens to land closer to a good IoU match on those particular layouts. Ours is ahead on 7 of 10 floors.


\begin{table}[t]
\caption{Room mIoU by floor, on the top-down grid (Sec.~\ref{sec:metrics}). $n_{\mathrm{pred}}$ is each method's predicted room count (HOV-SG / ours) against that floor's own observed region count. The final row is the mean over floors. \textbf{Bold}: better mIoU per floor. Ours is ahead on 7 of 10; the paired Wilcoxon signed-rank test over these differences gives $p = 0.065$, so the margin is reported as a trend.}
\label{tab:scenewise}
\centering
\small
\setlength{\tabcolsep}{5pt}
\setlength{\aboverulesep}{0pt}\setlength{\belowrulesep}{0pt}
\setlength{\extrarowheight}{0.4ex}
\begin{fittable}
\begin{tabular}{lccccc>{\columncolor{oursgrey}}c}
\toprule
Scene & Fl. & $n_{\mathrm{gt}}$ & \multicolumn{2}{c}{$n_{\mathrm{pred}}$} & \multicolumn{2}{c}{room mIoU$\uparrow$} \\
\cmidrule(lr){4-5}\cmidrule(l){6-7}
 & & & HOV-SG & Ours & HOV-SG & Ours \\
\midrule
\texttt{00824}-Dd4bFSTQ8gi & 0 & 10 & 8 & 11 & 0.816 & \textbf{0.887} \\
\texttt{00829}-QaLdnwvtxbs & 0 &  7 & 5 &  7 & \textbf{0.842} & 0.795 \\
\addlinespace[2pt]
\texttt{00843}-DYehNKdT76V & 0 &  7 & 3 &  5 & 0.611 & \textbf{0.734} \\
\texttt{00843}-DYehNKdT76V & 1 &  7 & 5 &  9 & 0.692 & \textbf{0.804} \\
\addlinespace[2pt]
\texttt{00873}-bxsVRursffK & 0 &  4 & 2 &  4 & 0.350 & \textbf{0.647} \\
\texttt{00873}-bxsVRursffK & 1 &  9 & 6 &  9 & 0.798 & \textbf{0.885} \\
\addlinespace[2pt]
\texttt{00877}-4ok3usBNeis & 0 &  7 & 3 &  8 & 0.698 & \textbf{0.859} \\
\texttt{00877}-4ok3usBNeis & 1 &  9 & 3 &  8 & \textbf{0.755} & 0.689 \\
\addlinespace[2pt]
\texttt{00890}-6s7QHgap2fW & 0 &  6 & 4 &  8 & \textbf{0.842} & 0.781 \\
\texttt{00890}-6s7QHgap2fW & 1 &  6 & 5 &  5 & 0.936 & \textbf{0.939} \\
\midrule
\multicolumn{5}{l}{\emph{mean over floors}} & 0.734 & \textbf{0.802} \\
\bottomrule
\end{tabular}
\end{fittable}
\end{table}

\section{Ablation Figures}
\label{app:ablation-figs}
\begin{figure}[!htbp]
\centering
\includegraphics[width=\columnwidth]{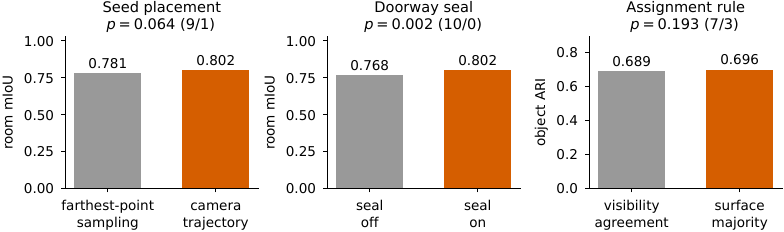}
\caption{The three ablations of Table~\ref{tab:ablation}, each scored on the metric it affects and
each with the variant used in the final method in orange: seed placement and the doorway seal on
room mIoU, the assignment rule on object ARI. Paired Wilcoxon $p$ and the win/loss split over the
ten floors are printed above each panel.}
\label{fig:ablation-sampling}
\end{figure}

Figure~\ref{fig:ablation-sampling} accompanies the ablation of Sec.~\ref{sec:res-ablation}.

\section{Further Experimental Material}
\label{app:deferred}

\subsection{Closure Scale}
\label{sec:res-scale}
Per-room closure-iteration is not persisted by the current pipeline, so no closure-scale distribution is reported here; the qualitative closure sequence of Figure~\ref{fig:pipeline} illustrates the effect.

\subsection{Runtime}
\label{sec:res-runtime}

Measured on an AMD Ryzen 7 5800H (8 cores, 16 threads) with 7\,GB of RAM, no GPU, single-threaded, over the ten evaluated floors at $0.05\,\mathrm{m}$ per cell. Reconstruction is excluded, since on HM3D it is replaced by the rendered walk (Sec.~\ref{sec:datasets}) and is shared by both methods.

Room detection takes $6.6 \pm 3.4$\,s per floor against $0.3 \pm 0.1$\,s for the HOV-SG baseline on the same map, so our method is roughly twenty times slower. The cost follows from the mechanism: the closure loop performs one dilation and one Euclidean distance transform over the $N \times N$ grid per iteration, giving $O(T N^{2})$ for an iteration cap $T$, where HOV-SG runs a single watershed. The remaining stages are minor by comparison: rasterisation $0.2$\,s, and the nearest-labelled-point completion applied to both methods $0.8$\,s. Runtime scales with the reconstruction rather than with the room count, ranging from $3.2$\,s on the smallest floor ($0.80$\,M points) to $13.4$\,s on the largest ($1.85$\,M).

Extracting door frames adds $5.5 \pm 1.9$\,s per floor, but this figure is dominated by loading and sampling the annotated mesh once per floor rather than by the hulling itself. In the monocular pipeline the frames arrive as a segmentation output already computed during reconstruction, so the equivalent cost is the projection and convex hull alone.

\end{document}